\documentclass{article}

\usepackage{microtype}
\usepackage{graphicx}
\usepackage{subfigure}
\usepackage{booktabs} 

\usepackage{hyperref}

\usepackage[accepted]{icml2023}

\usepackage{amsmath}
\usepackage{amssymb}
\usepackage{mathtools}
\usepackage{amsthm}

\usepackage[capitalize,noabbrev]{cleveref}

\theoremstyle{plain}
\newtheorem{theorem}{Theorem}[section]

\theoremstyle{definition}
\newtheorem{definition}[theorem]{Definition}

\theoremstyle{remark}

\usepackage[textsize=tiny]{todonotes}

\icmltitlerunning{Toward Unlimited Self-Learning MCMC with Parallel Adaptive Annealing}

\begin{document}

\twocolumn[
\icmltitle{Toward Unlimited Self-Learning MCMC with Parallel Adaptive Annealing}




\begin{icmlauthorlist}
\icmlauthor{Yuma Ichikawa}{yyy}
\icmlauthor{Akira Nakagawa}{yyy}
\icmlauthor{Hiromoto Masayuki}{yyy}
\icmlauthor{Yuhei Umeda}{yyy}
\end{icmlauthorlist}

\icmlaffiliation{yyy}{Fujitsu Limited}

\icmlcorrespondingauthor{Yuma Ichikawa}{ichikawa.yuma@fujitsu.com}

\icmlkeywords{Machine Learning, ICML}

\vskip 0.3in
]



\printAffiliationsAndNotice{}  

\begin{abstract}
Self-learning Monte Carlo (SLMC) methods are recently proposed to accelerate Markov chain Monte Carlo (MCMC) methods using a machine learning model. With latent generative models, SLMC methods realize efficient Monte Carlo updates with less autocorrelation. 
However, SLMC methods are difficult to directly apply to multimodal distributions for which training data are difficult to obtain. 
To solve the limitation, we propose ``parallel adaptive annealing,'' which makes SLMC methods directly apply to multimodal distributions with a gradually trained proposal while annealing target distribution. 
Parallel adaptive annealing is based on (i) sequential learning with annealing to inherit and update the model parameters, (ii) adaptive annealing to automatically detect under-learning, and (iii) parallel annealing to mitigate mode collapse of proposal models. We also propose VAE-SLMC method which utilizes a variational autoencoder (VAE) as a proposal of SLMC to make efficient parallel proposals independent of any previous state using recently clarified quantitative properties of VAE\@. Experiments validate that our method can proficiently obtain accurate samples from multiple multimodal toy distributions and practical multimodal posterior distributions, which is difficult to achieve with the existing SLMC methods. 
\end{abstract}

\section{INTRODUCTION}
High-dimensional probability distributions for which the normalizing factor is not analyzable appear in a wide variety of fields. For example, they appear in condensed-matter physics \citep{binder1993monte, baumgartner2012monte}, biochemistry \citep{manly2018randomization}, and Bayesian inference \citep{martin2011mcmcpack, gelman1995bayesian, foreman2013emcee}. Markov chain Monte Carlo (MCMC) methods are powerful and versatile numerical methods for sampling from such probability distributions. Its efficiency strongly depends on the choice of the proposal.

Among recent advances in machine learning, a general method called the self-learning Monte Carlo (SLMC) method \citep{liu2017self} was introduced to accelerate MCMC simulations by an automated proposal with a machine learning model and has been applied to various problems \citep{xu2017self, shen2018self}. 
In particular, a latent generative model realizes efficient global update through the obtained information-rich latent representation \citep{huang2017accelerated, albergo2019flow, monroe2022learning, tanaka2017towards}.
Although powerful, the performance of SLMC simulation strongly depends on an automated proposal with machine learning models and the quality of training data to train the proposal. 
For example, it is challenging to directly use SLMC for multimodal distributions because obtaining accurate training data covering all modes is difficult. 

In this paper, we propose a novel SLMC method to directly make the methods applicable to multimodal distributions and substantially expand the scope of application of the SLMC methods in various fields.
Specifically, we introduce (i) sequential learning with annealing to inherit and update the model parameters for pre-train effect, like a simulated annealing \citep{kirkpatrick1983optimization}, (ii) adaptive annealing path for efficient and automatic detection of the under learning, 
(iii) parallel annealing method mitigates that the proposal does not generate all modes of the target distribution, i.e., ``mode collapse'' by ensemble learning effect.
The resulting combination is called the ``parallel adaptive annealing SLMC''. 
With only a simple annealing (i), under-learning and mode collapse of proposals typically lead to failure results, but parallel adaptive annealing can solve these problems and even speed up. This improvement significantly contributes to using SLMC methods for larger systems.
Additionally, as an example of a proposal in this method, we propose a variational autoencoder (VAE) \cite{kingma2013auto}, one of the latent generative models, as a proposal by applying the recently clarified quantitative property of VAEs \citep{nakagawa2021quantitative, rolinek2019variational}.
Our contributions are summarized as follows:

\begin{itemize}
    \item Introduction of annealing for SLMC methods, which make SLMC methods directly applicable to multimodal distribution.
    \item SLMC-specific parallel adaptive annealing method, which automatically detect and solve under-learning and mode collapse in annealing and even speed up the annealing. 
    \item Through numerical experiments, we demonstrate that parallel annealing method can solve the under-learning and mode collapse and also apply the method to various difficult problems.
\end{itemize}

\section{BACKGROUND}\label{sec:background}
\subsection{Metropolis--Hastings Algorithm}\label{subsec:mcmc}

We start this section by illustrating the Metropolis--Hastings method \citep{hastings1970monte}. 
We define $p$ as a target distribution on a state space $\mathcal{X}$ where the normalizing factor is not necessarily tractable. 
Then assume  $\tilde{p} \propto p$ is tractable which can be expressed in closed form.
To obtain samples from $p$, we use an MCMC method to construct a Markov chain whose stationary distribution is $p$. The objective of the MCMC method is to obtain a sequence of correlated samples from the target distribution $p$, where the sequence can be used to compute an integral such as an expected value. 

The Metropolis--Hastings algorithm is a generic method to construct the Markov chain \citep{hastings1970monte}. The Metropolis--Hastings algorithm prepares a conditional distribution $q$ called a proposal distribution, which satisfies irreducibility, and the Markov chain $(x^{(t)})_{t=0}^{T}$ proceeds via the following steps: starting from $x^{(0)} \sim \pi_{0}$, where $\pi^{(0)}$ is the initial distribution, at each step $t$, proposing $x^{\prime} \sim q(x^{\prime}|x^{(t)})$, and with probability
\begin{equation*}
 A(x^{\prime}|x^{(t)}) = \min \left(1, \frac{\tilde{p}(x^{\prime}) q(x^{(t)}|x^{\prime})}{\tilde{p}(x_{t}) q(x^{\prime}|x^{(t)})}  \right),   
\end{equation*}
accept $x^{\prime}$ as the next update $x^{(t+1)} = x^{\prime}$ or with probability $1 - A(x^{\prime}|x^{(t)})$, reject $x^{\prime}$ and retain the previous state $x^{(t+1)} = x^{(t)}$. The proposal distribution $q(x^{\prime}|x^{(t)})$ is prepared for typically a local proposal, such as $q(x^{\prime}|x^{(t)}) = \mathcal{N}(x^{\prime}| x^{(t)}, \Sigma)$, where a scale parameter $\Sigma$ for the continuous probability distribution.
Hereafter, such a method is referred to as a local update MCMC method. Note that the calculation of the acceptance probability only requires the ratio of probability distributions (i.e., calculation of the normalizing factor is not required). Note that the performance of the algorithm strongly depends on the choice of the proposal distribution $q$. 

\subsection{Self-learning Monte Carlo}\label{subsec:slmc}
Motivated by the developments of machine learning, a general method, known as the self-learning Monte Carlo (SLMC) method, has been proposed. In this method, the information of samples generated by a local update MCMC method is extracted by a machine learning model $p_{\theta}(x)$ and the MCMC simulation is accelerated with the model $p_{\theta}(x)$. Specifically, the SLMC method sets the proposal distribution as $q(x^{\prime}|x^{(t)}) = p_{\theta}(x^{\prime})$ and the acceptance probability is expressed as
\begin{equation}
\label{eq:slmc-aprob}
    A(x^{\prime} | x^{(t)}) = \min \left(1, \frac{\tilde{p}(x^{\prime}) {p}_{\theta}(x^{(t)})}{\tilde{p}(x^{(t)}) {p}_{\theta}(x^{\prime})}\right).
\end{equation}
If we obtain a perfect machine learning model such that $p = p_{\theta}$, the acceptance probability is $1$. Moreover, if the proposal $x^{\prime} \sim p_{\theta}(x^{\prime})$ is accepted, the state will be uncorrelated with any previous state because the proposal probability $p_{\theta}(x^{\prime})$ is independent of the previous state $x^{(t)}$.

Note that the machine learning model $p_{\theta}(x)$ for the SLMC method is desired to have the following three properties: 
\begin{enumerate}
    \item The model should have the expressive power to well approximate the target probability distribution $p(x)$.
    \item Both proposals and acceptance probabilities in Eq.~\eqref{eq:slmc-aprob} should be computed from the machine learning model $p_{\theta}(x)$ at low cost.
    \item accurate training data should be available to train the machine learning model $p_{\theta}(x)$.
\end{enumerate}
If the first property is unsatisfied, the proposed state $x^{\prime}$ is rarely accepted and the correlation between samples becomes large. 
If the second property is unsatisfied, then, even if a good machine learning model $p_{\theta} \approx p$ is obtained, the SLMC simulation will not be efficient because of the cost of proposing a new state $x^{\prime}$ and computing the acceptance probability in Eq.~\eqref{eq:slmc-aprob}.
If the third property is not satisfied,
the SLMC method will generate biased samples in practical sense due to slow mixing, etc.
because the model can hardly propose a state that is not included in the training data.

\subsection{Theoretical results of VAE}\label{subsec:implict-isometricity}
A VAE \cite{kingma2013auto} is a latent generative model, and its purpose is to model the probability distribution behind the data. 
Let $\mathcal{D} = \{x^{\mu}\}_{\mu=1}^{P}$, $x^{\mu} \in \mathbb{R}^{D}$ be training data and $p_{\mathcal{D}}(x)$ be the empirical distribution of the training data. Specifically, the VAE is a probability model of the form $p_{\theta}(x, z) = p_{\theta}(x|z)p_{\theta}(z)$, where $z \in \mathbb{R}^{M}$ is the latent variable with prior $p_{\theta}(z)$, which is a standard normal distribution widely used. 
The $\beta$-VAE \citep{higgins2016beta} is trained by maximizing the following evidence lower bound (ELBO) of log-likelihood with variational posterior $q_{\phi}(z|x) = \mathcal{N}(z ; \mu_{\phi}(x), \mathrm{diag}(\sigma^{2}_{\phi}(x)))$, where $\mu_{\phi}(x) = (\mu_{m, \phi}(x))_{m=1}^{M}$, $\sigma_{\phi}(x) = (\sigma_{m, \phi}(x))_{m=1}^{M}$ is an estimating parameter, since direct maximization of the log-likelihood is difficult:
\begin{multline}
\label{eq:beta-elbo}
    \mathbb{E}_{p_{\mathcal{D}}} \left[\mathbb{E}_{q_{\phi}}[\log p_{\theta}(x| z)] - \beta_{\mathrm{VAE}} D_{\mathrm{KL}}[q_{\phi}(z|x) \| p(z)]  \right] \\
    \equiv \mathbb{E}_{p_{\mathcal{D}}}[\mathcal{L}(x, \theta, \phi;\beta_{\mathrm{VAE}})].
\end{multline}
where $D_{\mathrm{KL}}[\cdot \| \cdot]$ and $\mathcal{L}_{\mathrm{ELBO}}$ denote Kullback–Leibler divergence and ELBO, respectively. $\beta_{\mathrm{VAE}} \in \mathbb{R}_{+}$ is introduced to control the trade-off between the first and second terms of Eq.~\eqref{eq:beta-elbo}.
$p_{\theta}(x|z)$ and $q_{\phi}(z|x)$ are called a decoder and an encoder, respectively, and the first term in Eq.~\eqref{eq:beta-elbo} is called a reconstruction loss.

Recently, extensive work has been done to clarify the theoretical properties of VAEs. In that theoretical analysis, we apply the following theorem \cite{rolinek2019variational, nakagawa2021quantitative}; see Supplementary Material \ref{sec:proof-sketch-of-vae} for detail of proof sketch.

\begin{theorem}
\label{thm:isometry}
Suppose that VAE satisfy the ``polarized regime'', which the latent space $\mathcal{Z}$ can be partitioned as the disjoint union $\mathcal{Z}_{\mathrm{a}}$, $\mathcal{Z}_{\mathrm{p}}$ such that; (a) $\forall z_{m} \in \mathcal{Z}_{a}, \sigma_{m, \phi}^{2}(x) \ll 1$, (b) $\forall z_{m} \in \mathcal{Z}_{\mathrm{p}}, \mu_{m, \phi}(x) \approx 0, \sigma_{m, \phi}^{2}(x) \approx 1, \partial_{z_{m}} \mathrm{Dec}_{\theta}(z)=0$. If the reconstruction loss is the sum of squared errors (SSE) and the prior $p(z)$ is a standard normal distribution, then following holds:
\begin{align}
\label{eq:vae-likelihood}
    p_{\theta}(x) 
    &\propto \mathcal{N}(\mu_{\phi}(x); 0_{M}, I_{M}) \prod_{m=1}^{M} \sigma_{m, \phi}(x) \equiv \Gamma_{\theta}(x).
\end{align}
under the optimally conditions of loss function Eq.~\ref{eq:beta-elbo}. Here we denote $0_{M} = (0, \ldots, 0) \in \mathbb{R}^{M}$ and $I_M$ denotes an $M$ dimensional identity matrix.
\end{theorem}
Whether Eq.~\ref{eq:vae-likelihood} holds, which is the key to the detailed valance,  can be easily validated in the trained VAE model as shown in Supplementary Material~\ref{subsec:checking-isometricity}.
Polarized regime is typically observed very early in the training, which is well-known to practitioners and is checked on multiple tasks and datasets, and the optimally conditions are easily reachable since \citet{rolinek2019variational} shows that every local optimal solution is a global optimal solution. 
Here, $\mathcal{Z}_{\mathrm{a}}$ and $\mathcal{Z}_{\mathrm{p}}$ are regarded as subspaces of the data manifold which a VAE learned and out of the manifold, respectively.
Accordingly, $M$ can be less than $D$ and must be larger than the dimension of the data manifold the VAE model can learn.

\section{ANNEALING SELF-LEARNING MONTE CARLO WITH VAE}\label{sec:annealing-slmc}
We introduce an annealing for SLMC methods to apply the methods to multi-modal distributions directly and clarify that simple annealing has two practical problems (under-learning and mode collapse). Then, we propose adaptive annealing to automatically detect under-learning of the proposal and parallel annealing to mitigate mode collapse of the proposal. In addition, we introduce a novel SLMC method with a VAE called ``VAE-SLMC,'' by applying the recently revealed theoretical results Eq.~\ref{eq:vae-likelihood}.

\subsection{Annealing for SLMC methods} 
\label{sec:annealing}
We start by introducing inverse temperature $\beta$ for a range of values $0 \le \beta \le 1$ to a target distribution $p$ and $\tilde p \propto p$ as
\begin{equation}
    p(x;\beta) = \frac{p(x)^{\beta} }{\int p(x)^{\beta} dx},
    \ \tilde {p} (x;\beta) = {\tilde p(x)^{\beta}},
\end{equation}

Note that $p(x; \beta=1) = p(x)$ becomes a target distribution. Typically, smaller $\beta$ allows easier exploration for multiple modes, even in the complex distribution. Our annealing for SLMC methods is based on increasing $\beta$ from a small value (i.e., high temperature) to 1, like simulated annealing. 

\begin{figure}
    \centering
    \includegraphics[width=\columnwidth]{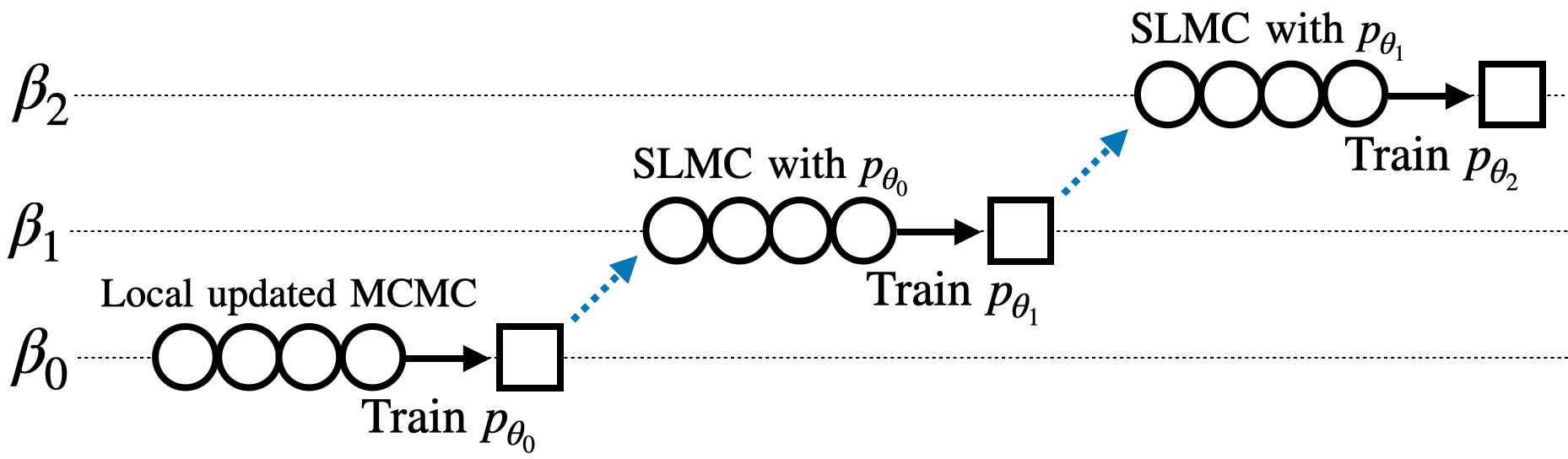}
    \caption{A graphical model of the annealing. 
    A circle denotes a state of a Markov chain with the stationary distribution $p(x;\beta_{k})$ and
    a square represents a state of the model.
    A black solid arrow means training of the model and a blue dashed arrow expresses the annealing procedure.}
    \label{fig:annealing-slmc-algorithm}
\end{figure}

The detailed flow of the annealing is abstracted in Fig.~\ref{fig:annealing-slmc-algorithm} and described following; First, we set up an annealing path $\{\beta_{k}\}_{k=0}^{K}$, $\beta_{0} \le \cdots  \le \beta_{K}=1$, where $\beta_{0}$ is set sufficiently small for the local update MCMC method to explore efficiently. Second, we acquire the training data from the $p(x;\beta_{0})$ by the local update MCMC method and train the model $p_{\theta_{0}}(x;\beta_{0})$ with the training data. Third, we acquire the training data again from the $p(x; \beta_{1})$ by the SLMC with the model $p_{\theta_{0}}(x; \beta_{0})$ and then train the model $p_{\theta_{1}}(x; \beta_{1})$ with the training data from $p(x; \beta_{1})$. Note that the initial parameters of training the model $p_{\theta_{1}}(x;\beta_{1})$ are set as $\theta_{0}$ since $p_{\theta_{0}}(x; \beta_{0})$ and $p_{\theta_{1}}(x; \beta_{1})$ are expected to have a similar common structure, leading to a similar effect like ``pre-training''. 
By repeating this process until $\beta_{K}=1$, we finally obtain accurate samples from the target distribution $p(x; \beta_{K}=1)$. 

Note that it is essential that the machine learning models not only satisfy the desired conditions explained in Sec.~\ref{subsec:slmc}, but also learn multimodal distributions and easily propose all learned modes to create accurate training data covering modes of target distributions in the annealing. Additionally, simple annealing (e.g., simply setting an arithmetic progression annealing path $\{\beta_{0} + k \Delta \beta\}_{k=0}^{K-1}$ and log-linear annealing path) often requires extra computation time and frequently lead to the following two problems, preventing SLMC methods from application to larger systems;
\begin{enumerate}
    \item \textbf{under-learning}: The acceptance rate decreases rapidly due to under-learning of the model while annealing.
    \item \textbf{mode collapse}: Although the acceptance rate is large, only partial support of a target distribution is proposed in practical time.
\end{enumerate}
In the following sections, we propose methods to solve these problems.

\subsection{Adaptive $\beta$ Scheduling for Annealing}\label{subsec:adaptive-annealing}
We propose a adaptive annealing, a method to adaptively determine the annealing path while automatically detecting ``under-learning'', following the two $\beta$-search strategies shown in Fig.~\ref{fig:adaptive-annealing-algorithm}. These two strategies can be used together as the algorithm in Supplementary Material \ref{subsec:aa-vae-slmc}.

\paragraph{Parallel search.} 
The candidates of $\{\beta_{(s)}^{\prime}\}_{s=0}^{S}$ are generated in parallel to determine $\beta_{k+1}$ from $\beta_{k}$.
The acceptance rates $\{\mathrm{AR}(\beta_{(s)}^{\prime})\}_{s=1}^{S}$ with respect to $\{p(x; \beta^{\prime}_{(s)})\}_{s=0}^{S}$ are then estimated in parallel by a typically small VAE-SLMC simulation with the model $p_{\theta_{k}}(x;\beta_{k})$, taking advantage of the fact that a VAE can make parallel proposals completely independent of any previous states.
After that, the $\beta_{(s)}^{\prime}$ that satisfies $\mathrm{AR}_{\min} \le \mathrm{AR}(\beta_{(s)}^{\prime}) \le \mathrm{AR}_{\max}$ is selected as a candidate of $\beta_{k+1}$.
Note that $\mathrm{AR}_{\min}$ and $\mathrm{AR}_{\max}$ are lower and upper bounds determined by a user.
If there exist multiple candidates, a single candidate is selected according to user-defined criteria (e.g., the smallest $\beta_{(s)}^{\prime}$ is selected).

\paragraph{Sequential search.}
First, a candidate $\beta^{\prime}_{(0)} \ge \beta_{k}$ is generated and the acceptance rate $\mathrm{AR}(\beta^{\prime}_{(0)})$ with respect to $p(x; \beta^{\prime}_{(0)})$ is calculated by a typically small SLMC simulation with the model $p_{\theta_{k}}(x ; \beta_{k})$.
Then, if $\mathrm{AR}(\beta^{\prime}_{(0)})$ is between $\mathrm{AR}_{\min}$ and $\mathrm{AR}_{\max}$,
the candidate $\beta^{\prime}_{(0)}$ is accepted as $\beta_{k+1}$,
otherwise, it is rejected and another candidate $\beta^{\prime}_{(1)}$ is generated by
\begin{equation*}
    \beta^{\prime}_{(1)} \leftarrow 
    \begin{cases}
    \beta^{\prime}_{(0)} + \varepsilon & (\mathrm{AR}(\beta^{\prime}_{(0)}) \ge \mathrm{AR}_{\max}) \\
    \beta^{\prime}_{(0)} - \varepsilon & (\mathrm{AR}(\beta^{\prime}_{(0)}) \le \mathrm{AR}_{\min})
    \end{cases},
\end{equation*}
where $\varepsilon$ is a positive constant determined by a user.
The new candidate $\beta^{\prime}_{(1)}$ is checked whether it is accepted or not in the same way as described above.
This process is repeated until the $s$-th candidate $\beta_{(s)}^{\prime}$ is accepted, and finally
$\beta_{(s)}^{\prime}$ is set to $\beta_{k+1}$.

Note that the proposed methods have an implicit function to detect under-learning of the model $p_{\theta_{k}}(x;\beta_{k})$ and retrain it.
If the model is not trained enough, the candidates $\beta_{(s)}^{\prime}$ are all rejected and $\beta_{k+1} = \beta^{\prime}_{(s)}$ remains equal to $\beta_{k}$, inducing ``retraining'' of the model with the same $\beta_{k}$.
This is a great advantage of the proposed method over simple scheduling.

\begin{figure}[tb]
    \centering
    \includegraphics[width=\columnwidth]{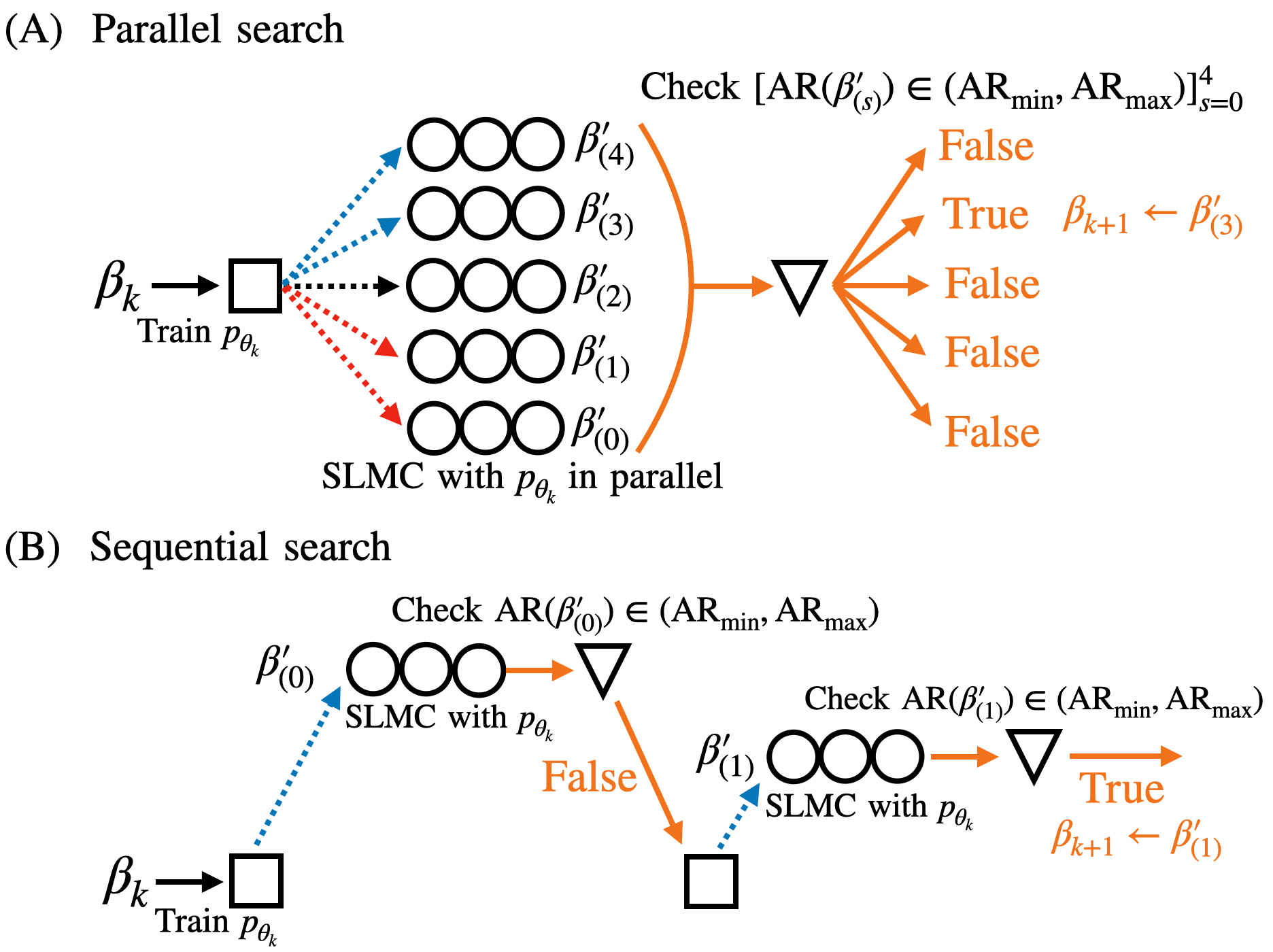}
    \caption{Graphical models illustrating the adaptive annealing.
    A triangle is a checking process for an acceptance rate,
    and the other notations are same as in Fig.~\ref{fig:annealing-slmc-algorithm}.
    If the check is True, output next $\beta_{k+1}$.}
    \label{fig:adaptive-annealing-algorithm}
\end{figure}

\subsection{Parallel Annealing}
\label{sec:parallel}
If the adaptive annealing includes the mode-collapsed model $p_{\theta_{k}}(x;\beta_{k})$ that does not train all modes of the target distribution, it generates samples not covering the all modes on steps larger than $\beta_{k}$.
We propose parallel annealing, which superimpose the states of each chain as in ensemble learning and to suppress mode collapse.

We first determine the number of parallel models and set up initial inverse temperature $\{\beta_{0}^{j}\}_{j=0}^{J}$, $\beta_{0}^{0}= \cdots = \beta_{0}^{J}$ where the upper/lower index represents the parallel chain/annealing step and then anneal each model in parallel as in the adaptive annealing. 
The only difference is that when creating the training data in the middle of the annealing, the states of the parallel chain are exchanged according to the following acceptance probabilities,
\begin{equation}
\label{eq:exchange-mc-prob}
    A(x_{k}^{j}, x_{k}^{j+1}) = \min \left(1, \frac{\tilde p(x_{k}^{j+1};\beta^{j}_{k}) \tilde p(x^{j}_{k};\beta_{k}^{j+1})}{\tilde p(x^{j}_{k};\beta_{k}^{j}) \tilde p(x_{k}^{j+1};\beta_{k}^{j+1})}\right),
\end{equation}
where is the same acceptance probabilities in the replica exchange Monte Carlo methods \cite{swendsen1986replica, hukushima1996exchange}. Note that our parallel annealing does not require careful adjustment of the beta intervals $\{\beta_{k}^{j}\}_{j=0}^{J}$ so that parallel chains with different inverse temperatures exchange well enough, as in the replica exchange Monte Carlo methods \cite{swendsen1986replica, hukushima1996exchange}. In parallel annealing, each model is first set to the same condition, $\beta_{0}^{0} = \cdots \beta_{0}^{J}$, so parallel chains exchange with probability 1, and an ensemble sample is generated from all chains as in ensemble learning. Then, the better models, with a high acceptance rate, have larger intervals from the worse models due to the adaptive annealing. As a result, the better model automatically exchanges with the other better models, and the poor models are rarely exchanged, yielding high-quality ensembled samples. Finally, there is no worry about considering the chains of worse models since we use the best model to reach $\beta=1.0$ most quickly.
As a result, even if the model $p_{\theta^{j}_{k}}(x^{j}_{k};\beta_{k}^{j})$ fails to learn entire region of the target distribution, the exchange process makes the training data for the model $p_{\theta^{j}_{k+1}}(x^{j}_{k+1};\beta_{k+1}^{j})$ the almost accurate samples. Also, even if there is no mode collapse, utilizing diverse models by the exchange is expected to produce high-quality ensembled training data, leading to better performance; see Supplementary Material \ref{subsec:aa-vae-slmc} for detailed implementation.

\begin{figure}
    \centering
    \includegraphics[width=\columnwidth]{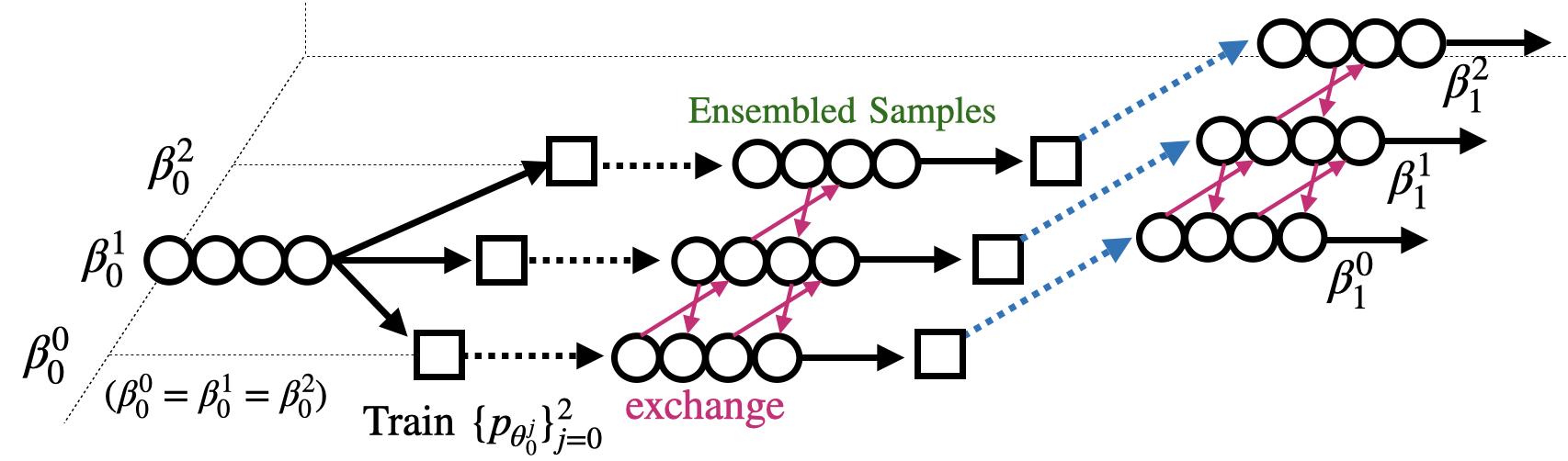}
    \caption{A graphical model illustrating the parallel annealing ESLMC method.
    Pink crossed arrows denote a exchange process between chains,
    and the other notations are same as in Fig.~\ref{fig:annealing-slmc-algorithm}.}
    \label{fig:parallel-annealing-algorithm.png}
\end{figure}

\subsection{Self-learning Monte Carlo with VAE}
Our VAE-SLMC method adopts a VAE as a proposal model ${p}_{\theta}(x)$ and
takes advantage of the VAE's theoretical properties to calculate acceptance probability efficiently.
We can easily compute $\Gamma_{\theta} \propto p_{\theta}$ from Eq.~\eqref{eq:vae-likelihood} by passing the state $x$ through the encoder $q_{\phi}(z|x)$, and the acceptance probability of the VAE-SLMC becomes
\begin{align}
\label{eq:ar-vae}
    &A(x^{\prime}| x^{(t)}) 
    = \min\! \left(1, \frac{\tilde{p}(x^{\prime}) \Gamma_{\theta}(x^{(t)})}{\tilde{p}(x^{(t)}) \Gamma_{\theta}(x^{\prime})}  \right). 
\end{align}
Note that detailed balance holds if Eq.~\eqref{eq:vae-likelihood} holds, and we validated not a concern in practice by efficient confirmability; see Supplementary Material \ref{subsec:checking-isometricity}.
The VAE can generate samples $x^{\prime} \sim p_{\theta}(x)$ at a low cost through the following procedure: The VAE first generates the latent variables, $z \sim \mathcal{N}(z; 0_{M}, I_{M})$ and then generates samples by passing the latent variables $z$ through the decoder $p_{\theta}(x| z)$. A well-trained VAE can make parallel proposals between modes independent of any previous state due to information-rich latent space. 
Moreover, we briefly confirmed that the method performs better than the methods of \cite{monroe2022learning} that also use VAE as a proposal and approximately evaluate the likelihood and of Flow-based model \cite{albergo2019flow} in Supplementary Material \ref{subsec:vae-vs-flow}. A detailed flow of the VAE-SLMC is shown in Algorithm~\ref{alg:VAE-SLMC}.
The model is updated by the ``retrain step'' with the samples generated by itself. 
Note that VAE-SLMC is applicable to discrete probability distributions. In addition, A detailed flow of the annealing VAE-SLMC is shown in Algorithm~\ref{alg:annealing-slmc}.

\begin{algorithm}[tb]
\caption{VAE-SLMC}\label{alg:VAE-SLMC}
\begin{algorithmic}
\renewcommand{\algorithmicrequire}{\textbf{Input:}}
 \renewcommand{\algorithmicensure}{\textbf{Output:}}
\REQUIRE $\tilde{p}(x)$, $p_{\theta}(x)$, $\Gamma_{\theta}(x)$; $T$, $T_{\mathrm{train}}$, $\mathcal{D}$,$\pi_{0}$ 
\STATE Initialize $x^{(0)} \sim \pi_{0}$
\WHILE{$t \le T$}
    \STATE Generate $x^{\prime} \sim p_{\theta}(x)$
    \STATE Calculate $A(x^{\prime} | x^{(t)})$ by Eq.~\eqref{eq:ar-vae}
\IF{($A(x^{\prime} | x^{(t)}) > U[0,1]$)}
    \STATE $x^{(t+1)} \gets x^{\prime}$ 
\ELSE
    \STATE $x^{(t+1)} \gets x^{(t)}$ 
\ENDIF
\IF{$t \equiv 0 ~(\mathrm{mod}~ T_{\mathrm{train}})$}
    \STATE Make training data $\mathcal{D}^{(t)}$ from $(x^{(t)})_{t=0}^{t}$ \\
    ($\rhd$ use nearly independent sequence of samples)
    \STATE Set $\mathcal{D} \gets \mathcal{D} \cup \mathcal{D}^{(t)}$ 
    \STATE Train $p_{\theta}(x)$ with $\mathcal{D}$ and update $\Gamma_{\theta}(x)$ 
\ENDIF
\STATE $t \gets t+1$
\ENDWHILE
\ENSURE samples $(x^{(t)})_{t=0}^{T}$
\end{algorithmic}
\end{algorithm}

\begin{algorithm*}
\caption{Annealing VAE-SLMC}\label{alg:annealing-slmc}
\begin{algorithmic}[1]
\renewcommand{\algorithmicrequire}{\textbf{Input:}}
 \renewcommand{\algorithmicensure}{\textbf{Output:}}
\REQUIRE target distribution $\tilde p(x;\beta)$, initial model $p_{\theta_{0}}(x ; \beta_{0})$, $\beta$-sequence $\{\beta_{k}\}_{k=0}^{K}$, initial distribution $\pi_{0}$, training data $\mathcal{D}$.
\STATE Initialize $x^{(0)} \sim \pi_{0}$
\FOR{$k=1$ to $K$}
    \STATE Set $(x^{(t)}_{k})_{t=0}^{T} = \text{VAE-SLMC}(\tilde p(x ; \beta_{k}), p_{\theta_{k-1}}(x ; \beta_{k-1}), \Gamma_{\theta_{k-1}}(x ; \beta_{k-1});T, T_{\mathrm{train}}, \pi_{0}, \mathcal{D}_{k-1})$ from Algorithm \ref{alg:VAE-SLMC}.
    \STATE Make the training data $\mathcal{D}_{k}$ from $(x^{(t)}_{k})_{t=0}^{T}$ 
    \STATE Train $p_{\theta_{k}}$ with $\mathcal{D}_{k}$
\ENDFOR
\STATE Set $(x^{(t)}_{K})_{t=0}^{T} = \text{VAE-SLMC}(\tilde p(x; \beta_{K}), p_{\theta_{K}}(x; \beta_{K}), \Gamma_{\theta_{K}}(x; \beta_{K}); T, T_{\mathrm{train}}, \pi_{0}, \mathcal{D}_{K})$
\ENSURE samples $(x^{(t)}_{K})_{t=0}^{T}$
\end{algorithmic}
\end{algorithm*}

\section{NUMERICAL EXPERIMENT}\label{sec:numerical-experiment}
We start by confirming the effectiveness of the parallel adaptive annealing and the performance on multi-cluster Gaussian mixture distributions. 
Then, we also validate the performance of the proposed method on real-world application to Bayesian inference of multimodal posterior for spectral analysis; see Supplementary Material \ref{sec:extended-explanation} for other discussions and practical applications.

\subsection{Experimental condition}
We summarize the essential settings for understanding the following results and discussions; 
the details of all the conditions of the numerical experiments are summarized in the Supplementary Material~\ref{sec:experiment-detail}. 

\paragraph{Abbreviation and proposals.}
Table~\ref{tab:ablation} summarizes the experimental conditions and their abbreviations. 
We experimented with three combinations of our proposed methods for VAE-SLMC, where VAE is set as $M=D$ for simplicity. We practically validated Eq.~\ref{eq:vae-likelihood} to ensure the detailed balance according to Supplementary Material~\ref{subsec:checking-isometricity}.
The parameters of MH, MH-EMC, and HMC-EMC are well-tuned. 
RW and HD in the proposal column of the table denote random walk proposal $ \mathcal{N}(x^{\prime} | x^{(t)}, \sigma^{2} I_{D})$ where $\sigma$ is scale factor and Hamilton dynamics \citep{duane1987hybrid}, respectively.

\paragraph{Initial learning.} 
For the CA-VAE-SLMC and AA-VAE-SLMC, we first train the VAE with 40,000 training data at $\beta_{0} = 0.1$, which are generated by a local update MCMC method. For each training data set, we trained VAEs until their losses nearly converged.

\begin{table}[tb]
\caption{Experimental Conditions and Abbreviations} \label{tab:ablation}
\begin{center}
\small
\begin{tabular}{lcccc}
Method & Anneal. & Adapt. & Parallel & Proposal \\
(Section) & (\ref{sec:annealing}) & (\ref{subsec:adaptive-annealing}) & (\ref{sec:parallel}) &  \\
\hline
CA-VAE-SLMC & $\surd$ & & & VAE  \\
AA-VAE-SLMC & $\surd$ & $\surd$ & & VAE  \\
AA-VAE-ESLMC & $\surd$ & $\surd$ & $\surd$ & VAE  \\
HMC-EMC & & & $\surd$ & HD  \\
MH-EMC & & & $\surd$ & RW  \\
MH & & & & RW  \\
\end{tabular}
\end{center}
\end{table}

\subsection{Gaussian Mixture}
We generalized the two Gaussian mixture proposed in \citep{woodard2009sufficient} to multi-Gaussian mixture. Specifically, each cluster $C=[1, 2, 3, 4, 5]$ is assumed to be Gaussian with variance $\sigma^{2}=0.5\sqrt{D/100}$ in the $D=[2, 10, 20, 25, 50, 100]$ dimension, the distance between each cluster is assumed to be equal, and the overall mean is assumed to be $0$.
Performance is measured via the root mean squared error (RMSE) defined as the Euclidean distance between the true expected value and its empirical estimate.

\paragraph{Effectiveness of parallel adaptive annealing.}
We compare three methods CA-VAE-SLMC, AA-VAE-SLMC and AA-VAE-ESLMC on three-cluster $20$D Gaussian mixture to confirm the effectiveness of parallel adaptive annealing. 

\begin{figure}[th]
    \centering
    \includegraphics[width=0.9\columnwidth]{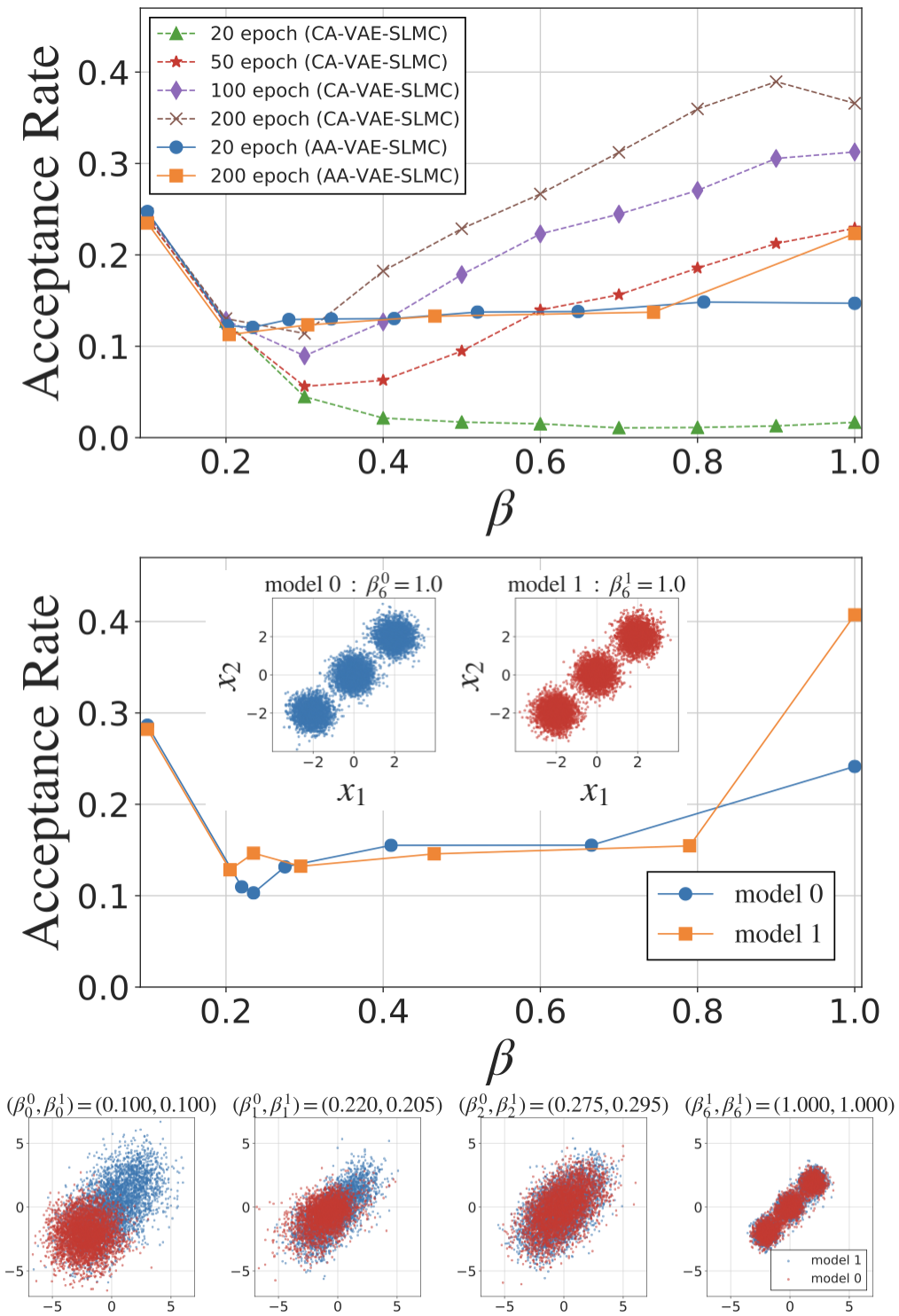}
    \caption{Effectiveness of parallel adaptive annealing. (a) Every dotted/solid line depicts the mean of acceptance rates of $10$ independent chains on annealing paths $\{\beta_{k}\}_{k=1}^{K}$. The marker/color represent the methods, CA-VAE-SLMC or AA-VAE-SLMC, and number of learning (epoch) of line~5 in Algorithm \ref{alg:annealing-slmc}. (b) Every dotted line also depicts the means of acceptance rate on each parallel annealing path, i.e, $\{\beta_{k}^{0}\}_{k=1}^{K}$ and $\{\beta_{k}^{1}\}_{k=1}^{K}$ on AA-VAE-SLMC. The color represent the models of AA-VAE-SLMC, model 0: initially mode-collapsed model and model 1: initially well-trained model. Inset figure show the samples generated AA-VAE-SLMC at $(\beta_{6}^{0}, \beta_{6}^{1})=(1.0, 1.0)$. (c) The samples generated by VAEs in the middle of parallel annealing on AA-VAE-SLMC.}
    \label{fig:effect-parallel-annealing}
\end{figure}

We start by confirming the effectiveness of adaptive annealing by comparing CA-VAE-SLMC and AA-VAE-SLMC, varying the number of learning, i.e., ``epoch'', in the middle of annealing (line 5 in Algorithm~\ref{alg:annealing-slmc}). 
The CA-VAE-SLMC uses linear annealing path, $\{\beta_{k}\}_{k=0}^{9} = \{0.1, 0.2, \ldots, 1.0\}$.
In AA-VAE-SLMC, the permissible acceptance rate is between $\mathrm{AR}_{\min}=0.1$ and $\mathrm{AR}_{\max} = 1.0$ in parallel search, and the largest of them is selected. Figure~\ref{fig:effect-parallel-annealing} (a) shows the relationship between the annealing paths and the acceptance rate corresponding to each beta in the annealing paths. We observe the pre-train effect that CA-VAE-SLMC with 50 epochs works, but for shorter learning, i.e., 20 epochs, the acceptance rate decreases rapidly due to the under-learning of a model. On the other hand, AA-VAE-SLMC with even 20 epochs automatically selects a narrow annealing interval in regions where learning is insufficient and prevents a sudden decrease in acceptance rate. Moreover, the total number of annealing times is 9, which is shorter than CC-VAE-SLMC. It's interesting to mention that AA-VAE-SLMC with 200 epochs greedily chooses a much shorter annealing path (6 times) due to sufficient learning. These result confirms that the adaptive annealing can automatically detect the under-learning and obtain more efficient annealing paths. Considering the pre-train effect, initially choosing a relatively small number of learning leads to reducing the total number of learning. In fact, the number of learning as a total during annealing of AA-VAE-SLMC with 20 epochs per training is 180 epochs, while AA-VAE-SLMC with 200 epochs per training is 1200 epochs. In addition, we observe that the acceptance rate can be controlled by just adjusting $\mathrm{AR}_{\min}$ and $\mathrm{AR}_{\max}$ in Supplementary Material~\ref{sec:dependency-lower-bound}.

Next, we confirm the effectiveness of parallel annealing. For this purpose, we prepared two VAEs of AA-VAE-ESLMC, one of which intentionally learn only one mode at $\beta_{0}^{0}=0.1$, and the other model that learned all modes at $\beta_{0}^{1}=0.1$; see $(\beta_{0}^{0}, \beta_{0}^{1}) = (0.1, 0.1)$ of Figure~\ref{fig:effect-parallel-annealing} (c).  
We observe that parallel annealing recover mode collapse, as shown in Figure~\ref{fig:effect-parallel-annealing} (c). Furthermore, Figure~\ref{fig:effect-parallel-annealing} (b) shows that adaptive annealing automatically reduces the annealing interval in the region where mode collapse begins to improve. Therefore, parallel annealing with at most two models recovers mode collapse well. The more well-learned models are combined, the earlier the mode collapse is recovered.

\begin{figure}[t]
    \centering
    \includegraphics[width=0.9\columnwidth]{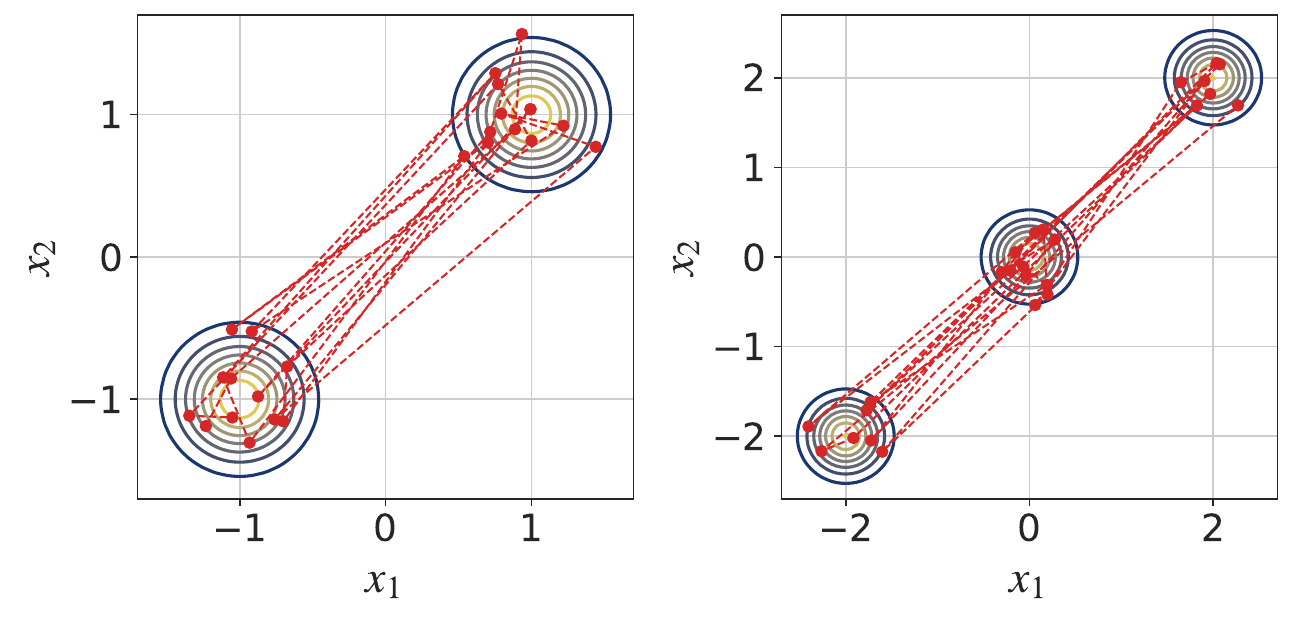}
    \caption{The transition of the AA-VAE-SLMC method on a two- or three Gaussian mixture. The contour lines represent the target distribution, the red dots are the sample sequence generated by the AA-VAE-SLMC, and the dotted lines represent the first 25 transitions.}
    \label{fig:adaptive-anneal-example}
\end{figure}

\begin{figure}[th]
    \centering
    \includegraphics[width=0.9\columnwidth]{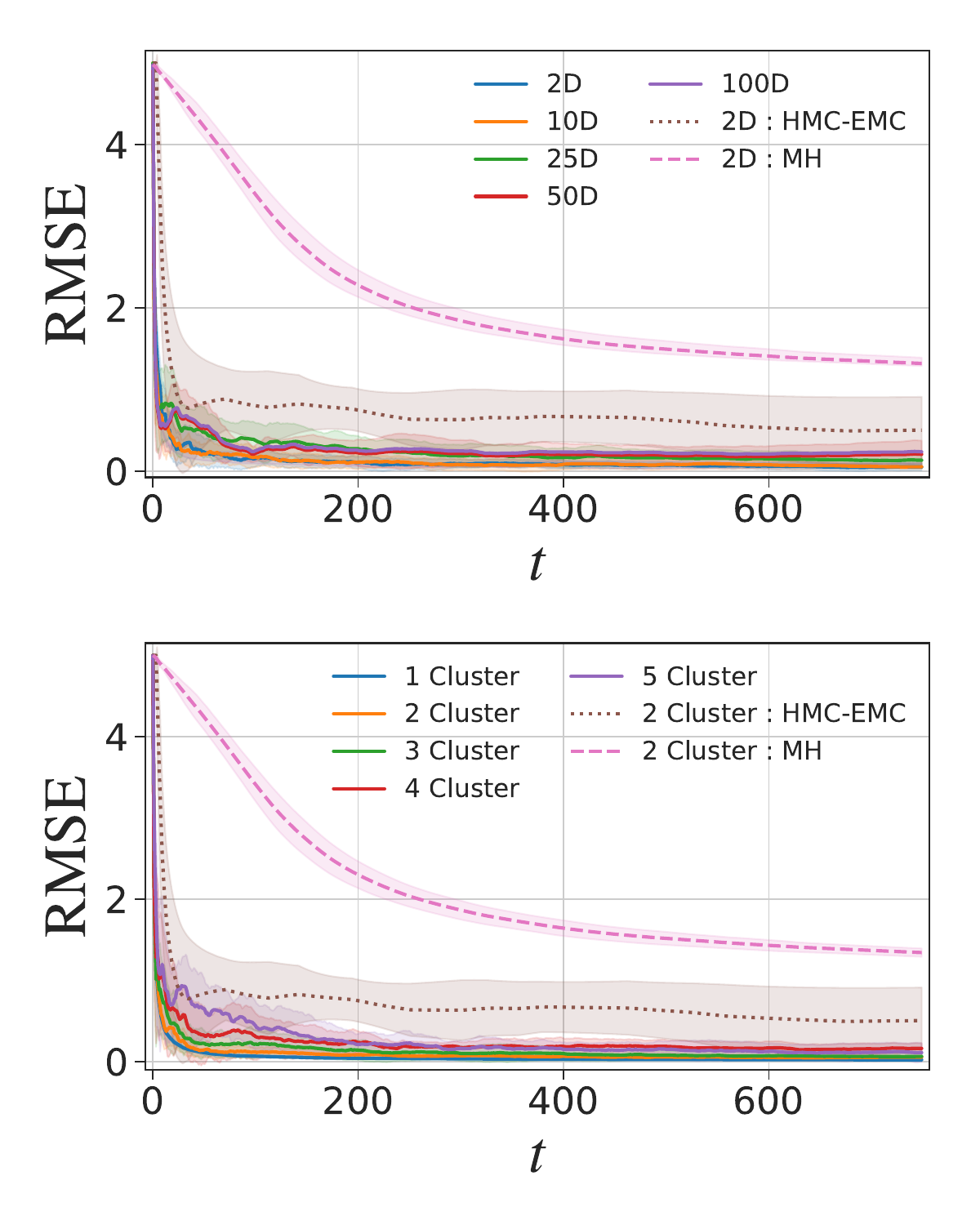}
    \caption{AA-VAE-SLMC method in Gaussian mixture. Every solid line/shaded area depicts the mean/standard deviation of RMSE averaged across $10$ independent chains after $t$-th Monte Carlo step. (Top) RMSE for Gaussian mixtures with fixed cluster $C=3$ and varying dimension $D$. (Bottom) RMSE for Gaussian mixtures with a fixed $D=10$ and varying cluster $M$.}    
\label{fig:adaptive-anneal-RMSEs}
\end{figure}

\paragraph{Results of adaptive annealing VAE-SLMC.}
Figure~\ref{fig:adaptive-anneal-example} shows the dynamic transitions between the modes by AA-VAE-SLMC with 200 epochs for a two- or three- cluster $2$D Gaussian mixture. 
Due to the information-rich latent space of the VAE,
the proposed method successfully realizes almost equally transitions between clusters at the early stage of the simulation, which is not possible for the local update MCMC method. 
Figure~\ref{fig:adaptive-anneal-RMSEs} shows the results when either $K$ or $D$ is varied and the other is fixed ($D=10$ or $K=3$). We can see that our AA-VAE-SLMC works for these multimodal distributions since the RMSEs drastically decrease compared with MH and HMC-EMC.

\subsection{Spectral Analysis}
\begin{figure}[th]
    \centering
    \includegraphics[width=0.9\columnwidth]{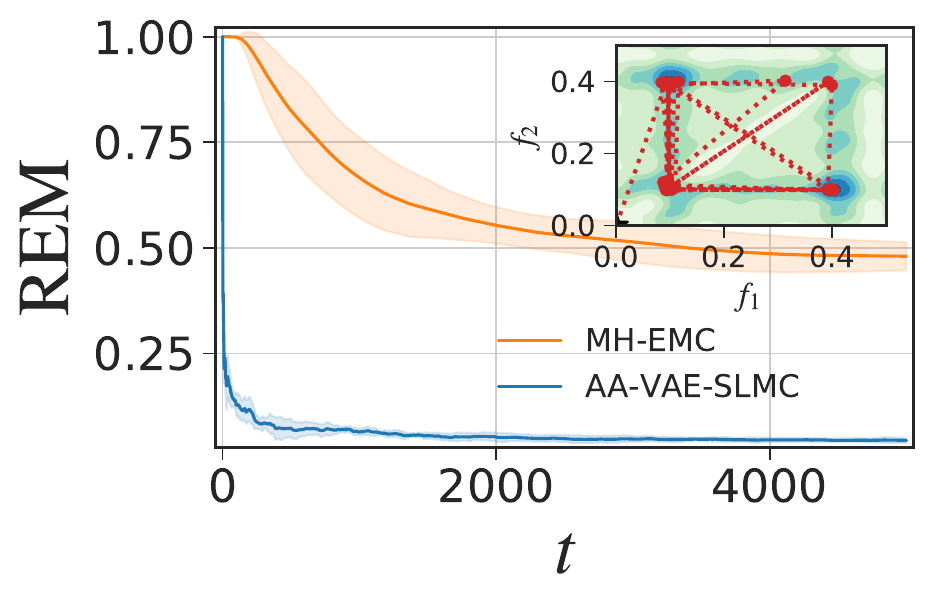}
    \caption{The relative error of the estimated mean (REM) for spectral analysis after $t$-th Monte Carlo step. Every solid line/shaded area depicts the mean/standard deviation of REM averaged across $10$ independent chains. In the inset figure, the contour line represents the marginal posterior, and the red dotted line represents the first $30$ transitions.}
    \label{fig:spectral-rem}
\end{figure}

Motivated by \citep{luengo2020survey}, we demonstrate real-world Bayesian estimation of a noisy multi-sinusoidal signal (see Supplementary Material \ref{subsec:sensor-problem} and \ref{subsec:optimization-problem} for the additional experiments of real-world Bayesian estimation of sensor network localization problems and optimization problems with multiple optimal solutions). The signal is given by
\begin{equation*}
    y(\tau) = A_{0} + \sum_{d=1}^{D} A_{d} \cos (2 \pi f_{d} \tau + \phi_{d}) + r(\tau), ~~\tau \in \mathbb{R},
\end{equation*}
where $A_{0}$ is a constant term, $A \equiv \{A_{d}\}_{d=1}^{D}$ is a set of amplitudes, $f \equiv \{f_{d}\}_{d=1}^{D}$ are frequencies, $\phi \equiv \{\phi_{d}\}_{d=1}^{D}$ are their phases, and $r(\tau)$ is an additive white Gaussian noise.
The estimation of the parameters is important in a variety of applications, such as signal processing \citep{stoica1993list, so2005linear}. 
Here we compute the posterior for $f \in \mathbb{R}^{8}$ given the data by discretizing $y(\tau)$. Note that the problem is symmetric with respect to hyperplaine $f_{1} = \cdots = f_{D}$ and the marginal posterior is multimodal (see Fig.~\ref{fig:spectral-rem}). 
Performance is measured via the relative error of the estimated mean (REM). We apply AA-VAE-SLMC starting from $\beta_{0}=0.1$, setting $\mathrm{AR}_{\min}=0.15$ and $\mathrm{AR}_{\max}=1.0$.
Figure~\ref{fig:spectral-rem} shows our method works for the problem since the REM is rapidly decreased compared with MH-EMC and all modes are covered in the first few transitions, compared with MH-EMC method.

\section{RELATED WORK}\label{sec:related-work}
SLMC methods with a latent generative model have been proposed. 
First, an SLMC with the restricted Boltzmann machine (RBM) \citep{huang2017accelerated} has been proposed, and the RBM resulted in an efficient global transition. The proposal of the RBM requires the MCMC simulation to be implemented by parallel alternating Gibbs sampling of hidden and visible variables. However, this method can cause problems related to long correlation times \citep{roussel2021barriers}. An SLMC with a Flow-based model has also been considered \citep{albergo2019flow}. The SLMC does not require MCMC and needs to worry about mixing. However, the Flow-based model requires strong restrictions for the networks. Similar to our method is the SLMC with a VAE proposed by \citep{monroe2022learning}. This method does not require MCMC and has no restriction on the network; however, this method is based on an approximation $p_{\theta}(z | x) \approx q_{\phi}(z | x)$, whose approximation can lead to a decrease in the acceptance rate, where evaluating the validity of the approximation at a low cost is difficult. Our proposed method can eliminate this approximation by applying the recently derived theoretical analysis of a VAE. They do not provide specific algorithms for situations where obtaining training data is challenging, such as multimodal distributions.

\section{CONCLUSION}
We have presented parallel adaptive annealing that makes SLMC methods applicable to situations where obtaining accurate training data is difficult. The annealing is based on (i) sequential learning with annealing target distribution to inherit and update the model, (ii) adaptive annealing to automatically detect under-learning, and (iii) parallel annealing to mitigate mode collapse of proposal models. We validate that this combination works on multimodal distribution while automatically detecting and recovering under-learning and mode collapse in the middle of annealing. We also propose the VAE-SLMC method, which utilizes a variational autoencoder (VAE) as the proposal to make efficient parallel proposals by the recently clarified properties of VAE\@. 

We believe that the annealing SLMC methods broaden the applications of the SLMC methods in various fields and the ability to automatically detect under-learning and mode collapse plays an important role in applications of SLMC methods to larger proposal machine learning models and more significantly complicated target distributions.
We also expect that continually evolving methods in hardware, software for deep neural network performance and machine learning techniques, including transfer learning and fine-tuning from a various pre-trained models, bring future practical gains for this framework. 

To the best of our knowledge, we are the first to consider specific annealing to SLMC methods for multimodal distributions while automatically detecting and recovering under-learning and mode collapse in the middle of annealing. In the future, it will be interesting to see the combination (e.g., small word proposal \citep{guan2006markov}) of gradient-based sampling methods such as variants of HMC \citep{duane1987hybrid} and Langevin MC Monte Carlo \citep{roberts2002langevin} dynamics and our parallel adaptive annealing SLMC methods with latent generative models for much higher dimensional and complex problems.



\bibliography{ref}
\bibliographystyle{icml2023}

\newpage
\appendix
\onecolumn

\setcounter{equation}{8}
\setcounter{algorithm}{2}
\setcounter{table}{1}
\setcounter{figure}{8}

\section{OVERVIEW}
This supplementary material provides extended explanations, implementation details, and additional results for the paper ``Toward Unlimited Self-Learning MCMC with Parallel Adaptive Annealing''.

\section{PROOF SKETCH}\label{sec:proof-sketch-of-vae}
In this section, we show a brief sketch of Eq.~\ref{eq:vae-likelihood} derivation in main text as a summary of previous works from \citet{rolinek2019variational, nakagawa2021quantitative, kato2020radogaga}. 
Here, we assume that the dimension of the data manifold is $N$, and this manifold is embedded in the $D$-dimensional input space.

First, we explain the case of $N=D$, i.e., the dimension of the data manifold is the same as the dimension of the input space for simplicity. 
In this case, we need to set $M=D$.
Here, \citet{kato2020radogaga} show that the loss function $\mathcal{L}(x, \theta, \phi;\beta_{\mathrm{VAE}})$ in Eq.~(\ref{eq:beta-elbo}) can be approximated as follows:
\begin{align}
\label{eq:expanded-lossa}
    \mathcal{L}(x, \theta, \phi;\beta_{\mathrm{VAE}}) &\approx \sum_{m=1}^M \sigma_{\phi, m}^{2}(x) \left(\frac{\partial x}{\partial \mu_{\phi}(x)} \right) G_{x} \left(\frac{\partial x}{\partial \mu_{\phi}(x)}  \right) 
- {\beta_{\mathrm{VAE}}} \log \left(p(x) \left|\mathrm{det}\left(\frac{\partial x}{\partial \mu_{\phi}(x)}  \right)  \right| \prod_{m=1}^M \sigma_{\phi, m}(x) \right),
\end{align}
where $G_{x}$ denote a metric tensor from $p_{\theta}(x | z)$ if $\forall m,~~\sigma_{m, \phi}^{2}(x) \ll 1$ holds. 
\citet{kato2020radogaga} further derived the optimal condition of Eq.~(\ref{eq:expanded-lossa}) as follows:
\begin{align}
\label{supeq:isoa}
\frac{2 \sigma^{2}_{m, \phi}(x)}{\beta_{\mathrm{VAE}}}\left(\frac{\partial x}{\partial \mu_{m, \phi}(x)} \right)^{\top} 
G_x
\left(\frac{\partial x}{\partial \mu_{n, \phi}(x)} \right) =  \delta_{mn},~~\forall m, n \in \{1, \ldots, M\},
\end{align}
where $\delta_{mn}$ denotes  the Kronecker delta. 
They called  the quantitative orthogonal property of Eq.~(\ref{supeq:isoa}) {\it implicit isometricity}.
In the case $p_{\theta}(x| z) = \mathcal{N}(x; \mu_{\theta}(z), \frac{1}{2} I_{D})$, the reconstruction loss becomes the sum of squared errors (SSE) and the metric tensor $G_{x}$ becomes an identity matrix $I_{D}$.  
From Eq.~(\ref{supeq:isoa}), we can derive the Jacobian determinant between $x$ and $\mu_{\phi}(x)$ as follows:
\begin{align}
\label{supeq:Jacob}
\left|\mathrm{det}\left(\frac{\partial \mu_{\phi}(x)}{\partial x} \right)\right| = \left(\frac{\beta_{\mathrm{VAE}}}{2} \right) ^{-{M}/{2}} \prod_{m=1}^M \sigma_{m, \phi}(x) \propto \prod_{m=1}^M \sigma_{m, \phi}(x)
\end{align}
Thus the probability of sample $p_\theta(x)$ can be derived from the prior and Jacobian determinant as follows: 
\begin{align}
\label{supeq:Prob}
p_\theta(x) = p(\mu_{\phi}(x)) \left|\mathrm{det}\left(\frac{\partial \mu_{\phi}(x)}{\partial x} \right)\right|  = p(\mu_{\phi}(x)) \  \left(\frac{\beta_{\mathrm{VAE}}}{2} \right) ^{-{M}/{2}} \prod_{m=1}^M \sigma_{m, \phi}(x) \propto p(\mu_{\phi}(x)) \prod_{m=1}^M \sigma_{m, \phi}(x)
\end{align}
Finally Eq.~(\ref{eq:vae-likelihood}) is derived by applying $p(\mu_{\phi}(x))=\mathcal{N}(\mu_{\phi}(x); 0, I_M)$ to Eq.~(\ref{supeq:Prob}).  

Second,  we explain the case of $N<D$, i.e., the dimension of the data manifold is less than the dimension of the input space. 
In this case, the value of the probability density cannot be defined in the $D$-dimensional input space.
Instead, we need to measure the probability density in the manifold space with the dimension $N$.
The following show an example.
Assume a 2-dimensional square plane manifold with edge length L in 3-dimensional space.
In the 3-dimensional space, the probability density of the point on the square plane manifold is infinite, and the probability density of the point out of the square plane manifold is zero.
By contrast, if we focus on the 2-dimensional manifold only, we can derive the probability density of the points on the square plane manifold as $1/L^2$.

In this case, \citet{rolinek2019variational} show that the following ``polarized regime'' is typically observed when a VAE is trained:
\begin{definition}
\label{def:polarized regime}
We say that parameters $\phi$, $\theta$ induce a ``polarized regime'' if the latent space $\mathcal{Z}$ can be partitioned to $\mathcal{Z}_{\mathrm{a}} $ and $\mathcal{Z}_{\mathrm{p}}$ (active and passive set) such that
\begin{itemize}
    \item[(a)] $\forall z_{m} \in \mathcal{Z}_{\mathrm{a}}$, $\sigma_{m, \phi}^{2} \ll 1$, 
    \item[(b)] $\forall z_{m} \in \mathcal{Z}_{\mathrm{p}}$, $\mu_{m, \phi}(x) \approx 0$,~~$\sigma_{m, \phi}^{2} \approx 1$, ~~$\partial_{z_{m}} \mathrm{Dec}_{\theta}(z) = 0$.
\end{itemize}
\end{definition}

\citet{rolinek2019variational} showed that polarized regime is typically observed very early in the training in the multiple data sets such as dSprites, MNIST, and Fashion MNIST.  \citet{nakagawa2021quantitative} also reported that the polarized regime is observed in CelebA dataset.
Here, $\mathcal{Z}_{\mathrm{a}}$ and $\mathcal{Z}_{\mathrm{p}}$ are regarded as subspaces of the data manifold which a VAE learned and  out of the manifold, respectively. 
Thus, 
Eq.~(\ref{eq:vae-likelihood}) can be expanded using \textit{polarized regime} (b) under the optimal conditions as follows:  
\begin{eqnarray}
\label{supeq:ProbDegen}
    p_{\theta}(x) 
     &\propto&
     \prod_{m | z_{m} \in \mathcal{Z}_{\mathrm{a}}} \mathcal{N}(\mu_{m, \phi}; 0, 1) \sigma_{m, \phi}(x) \ 
     \prod_{m | z_{m} \in \mathcal{Z}_{\mathrm{p}}} \mathcal{N}(\mu_{m, \phi}; 0, 1) \sigma_{m, \phi}(x)
     \nonumber \\
     &\approx&
    \prod_{m | z_{m} \in \mathcal{Z}_{\mathrm{a}}} \mathcal{N}(\mu_{m, \phi}; 0, 1) \sigma_{m, \phi}(x) \ (2 \pi)^{-{\mathrm{dim}(\mathcal{Z}_{p})}/{2}}
     \nonumber \\
     &\propto&
         \prod_{m | z_{m} \in \mathcal{Z}_{\mathrm{a}}} \mathcal{N}(\mu_{m, \phi}; 0, 1) \sigma_{m, \phi}(x)
,
\end{eqnarray}
since the second product term regarding $\mathcal{Z}_{\mathrm{p}}$ in the second equation is constant. 
Here, $p_{\theta}(x)$ in Eq.~(\ref{supeq:ProbDegen})  can be considered to be proportional to the probability density on the data manifold space $\mathcal{Z}_{\mathrm{a}}$ from the last equation of Eq.~(\ref{supeq:ProbDegen}). 
Thus, $M$, i.e., the dimension of $\mathcal{Z}$, can be be reduced down to $\mathrm{dim}(\mathcal{Z}_{\mathrm{a}})$, which will be corresponding to $N$.

Note that \citet{kato2020radogaga} introduced an isometric variational autoencoder called RaDOGAGA and showed the precise and mathematical derivation of the probability density when $N<D$.
Since the work of \citet{nakagawa2021quantitative} which show Eqs.~(\ref{eq:expanded-lossa})-(\ref{supeq:Prob}) is an expansion of \citet{kato2020radogaga}, Eq.~\ref{supeq:ProbDegen} can be mathematically derived in the same manner.

\section{EXTENDED EXPLANATIONS OF VAE's IMPLICIT ISOMETRICITY}\label{sec:extended-explanation}
In this section, we show how to check the consistency of the VAE's implicit isometricity shown in the paper and how to efficiently analyze the properties of probability distributions from the obtained samples using the implicit isometricity.

\subsection{Checking Isometricity of the VAE}\label{subsec:checking-isometricity}

This section shows how to check whether a VAE acquires the isometricity.
In the following discussion, we assume that the SSE metric is used as a reconstruction loss of the VAE objective. 
From Eq.~\eqref{supeq:isoa} and $G_x=I_M$, a following property
\begin{align}
\label{supeq:isofact}
\sqrt{\frac{2 \sigma^{2}_{m, \phi}(x)}{\beta_{\mathrm{VAE}}}}
\left \|\frac{\partial x}{\partial \mu_{m, \phi}(x)} \right \|_2 = 1
\end{align}
holds in each latent dimension $m$. Eq.~\eqref{supeq:isofact} can be numerically estimated as
\begin{align}
\label{supeq:isonum}
\sqrt{\frac{2 \sigma_{m, \phi}^{2}(x)}{\beta_{\mathrm{VAE}}}} \ 
     \left \|\frac{\partial x}{\partial \mu_{m, \phi}(x)} \right \|_2 
     \simeq 
     \sqrt{\frac{2 \sigma_{m, \phi}^{2}(x)}{\beta_{\mathrm{VAE}}}}
     \frac{\| \mathrm{Dec}_\theta( \mu_{\phi}(x)) - \mathrm{Dec}_\theta (\mu_{\phi}(x)+\Delta \  e_{(m)})\|_2}{\Delta},
\end{align}
where $\mathrm{Dec}_\theta : \mathbb{R}^{M} \to \mathbb{R}^{D}$ is a trained VAE decoder,
$\Delta$ is an infinitesimal constant for numerical differentiation,
and $e_{(m)}$ is a one-hot vector $(0,\cdots,1,\cdots,0)$,
in which only the $m$-th dimensional component is $1$ and the others are $0$.
We define the isometric factor for each latent dimension as 
\begin{equation}
\label{eq:IsoF}
\mathrm{Iso}_{m} = \mathbb{E}_{p_{\hat{\mathcal{D}}}}\left[\sqrt{\frac{2 \sigma_{m, \phi}^{2}(x)}{\beta_{\mathrm{VAE}}}}
     \frac{\| \mathrm{Dec}_\theta( \mu_{\phi}(x)) - \mathrm{Dec}_\theta (\mu_{\phi}(x)+\Delta \  e_{(m)})\|_2}{\Delta}  \right],
\end{equation}
where $\mathbb{E}_{p_{\hat{\mathcal{D}}}}$ denote the average value over the generated 100,000 samples from $p_{\theta}$.
If the isometric factor is close to $1$, the probability $p_{\theta}(x)$ can be correctly estimated by using  Eq.~(\ref{eq:vae-likelihood}) in the main document.
Note that the orthogonality holds as shown in 
 \citep{nakagawa2021quantitative, rolinek2019variational}.

Table~\ref{tab:ablation} shows the isometric factor $\mathrm{Iso}_{m}$ of each dimension $m$ of the VAE latent variable in a descending order of the importance calculated by Eq.~\eqref{eq:importance-vae}.
$\Delta$ in Eq.~\eqref{eq:IsoF} is set to $0.001$.
GMM-$[2,3,4,5]$ show Gaussian Mixtures ($D=10, M=10$) with $[2,3,4,5]$ clusters used in Sec.~4.1 in the main text.
ICG is Ill-Conditional Gaussian  ($D=100, M=100$) explained in Section~D.2. For ICG, top 10-dimensional values from 100 dimensions are listed in the table.
BANANA  ($D=2, M=2$), SCG  ($D=2, M=2$), and RW  ($D=2, M=2$) denote banana-shaped density,  strongly correlated Gaussian, and rough well distributions, respectively, as explained in Sec.~D.2. See Sec.~\ref{sec:experiment-detail} for details of the experiment conditions.
As shown in this table, the isometric factors for each dimension in our experiments are very close to 1.0, showing the proposal probability can be correctly estimated from  Eq.~(\ref{eq:vae-likelihood}) in the main document.

\begin{table}[tb]
\caption{Evaluation of the isometric factors $\mathrm{Iso}_{m}$ in the trained VAE models} \label{tab:ablation}
\begin{center}
\begin{tabular}{ccccccccc}
Latent dim. $m$ & GMM-2 & GMM-3 & GMM-4 & GMM-5 & ICG & BANANA & SCG & RW \\
1 & 0.974 & 0.978 & 0.996 & 0.987 & 1.009 & 1.009  & 1.000 & 0.999 \\
2 & 0.986 & 0.987 & 0.990 & 0.990 & 1.013 & 0.993  & 1.003 & 1.000 \\
3 & 0.990 & 0.987 & 0.990 & 0.992 & 1.013 &       &       &       \\
4 & 0.987 & 0.991 & 0.999 & 0.994 & 1.012 &        &       &       \\
5 & 0.993 & 0.993 & 0.992 & 0.996 & 1.000 &        &       &       \\
6 & 0.992 & 0.993 & 1.000 & 0.992 & 1.019 &        &       &       \\
7 & 0.992 & 0.995 & 0.996 & 0.991 & 1.014 &        &       &       \\
8 & 1.001 & 0.992 & 1.000 & 0.990 & 1.013 &        &       &       \\
9 & 0.999 & 1.004 & 1.001 & 0.995 & 1.004 &        &       &       \\
10 & 1.002 & 1.004 & 0.993 & 0.997 & 1.025 &        &       &       \\
\end{tabular}
\end{center}
\end{table}

\subsection{Understanding the Structure of Probability Distributions}
By using the properties of VAE, the structure of probability distributions can be analyzed at low cost. In Bayesian statistics and other probabilistic inference, principal component analysis (PCA) \citep{shlens2014tutorial} of the samples generated by a sampling method such as MCMC methods is sometimes performed to understand the structure of probability distribution \citep{inoue2005pca, inoue2006analysis, wang2016discovering, kiwata2019deriving}. However, the computational cost for calculating eigenvalues in PCA is $\mathcal{O}(\min(D^{3}, P^{3}))$, where $D$ is the data dimension and $P$ is the number of samples generated by an MCMC method. However, by using the VAE obtained by VAE-SLMC, a low-dimensional compression similar to PCA is possible just by encoding and decoding the samples with a cost of $\mathcal{O}(\Lambda D)$, where $\Lambda$ is a computational cost of a VAE's encoder.
The importance of each latent dimension $m=1, \ldots, M$ in the latent space can be evaluated \citep{nakagawa2021quantitative} as
\begin{equation}
\label{eq:importance-vae}
    \kappa_{m} = \frac{\beta}{2} \mathbb{E}_{p_{\theta}}\left[1/\sigma_{m, \phi}^{2}(x) \right]. 
\end{equation}
We substitute $\mathbb{E}_{p_{\theta}}$ for the empirical mean of $P$ samples and encoded variance $\sigma_{\phi}(x) = (\sigma_{m, \phi})_{m=1}^{M}$ can be obtained by passing $P$ samples through encoder-decoder. 

\section{ALGORITHM  DETAILS}
In this section, we describe the implementation details of the proposed VAE-SLMC with adaptive annealing and parallel annealing methods.

\subsection{Summary of the Proposed VAE-SLMC Methods}
Here, we summarize the variation of the proposed VAE-ALMC methods.
Table~\ref{tab:ablation-alg} lists all the combinations of the proposed method with
the adaptive annealing technique (in Sec.~3.3 and Algorithms~\ref{alg:adaptive-annealing-slmc-adaptive-beta} and~\ref{alg:adaptive-anneal})
and the parallel annealing exchange technique (in Sec.~3.4 and Algorithms~\ref{alg:parallel-anneal} and~\ref{alg:exchange-slmc}).
The abbreviations of the algorithms (CA-VAE-SLMC, AA-VAE-SLMC, and AA-VAE-ESLMC) used in the experiments are also described.
``CA-'' means constant annealing, which is a method without adaptive annealing.

\begin{table}[tb]
\caption{Experimental Conditions, Abbreviations, and Algorithms used in each Conditions} \label{tab:ablation-alg}
\begin{center}
\small
\begin{tabular}{lcccc}
& VAE-SLMC & Annealing & Adaptive annealing & Parallel annealing exchange \\
& (Sec. 3.1) &(Sec. 3.2) & (Sec. 3.3) & (Sec. 3.4)  \\
Method & (Alg. 1)& (Alg. 2) &  (Alg.~\ref{alg:adaptive-annealing-slmc-adaptive-beta} and \ref{alg:adaptive-anneal}) &  (Alg. \ref{alg:parallel-anneal} and \ref{alg:exchange-slmc})  \\
\hline
Na\"ive VAE-SLMC & $\surd$ & & & \\
CA-VAE-SLMC & $\surd$ & $\surd$ & & \\
AA-VAE-SLMC & $\surd$ & $\surd$ & $\surd$ & \\
AA-VAE-ESLMC & $\surd$ & $\surd$ & $\surd$ & $\surd$ \\
\end{tabular}
\end{center}
\end{table}

\subsection{Adaptive Annealing VAE-SLMC}
\label{subsec:aa-vae-slmc}
Algorithm~\ref{alg:adaptive-annealing-slmc} illustrates the steps of adaptive annealing VAE-SLMC (AA-VAE-SLMC) explained in Sec.~3.3 of the main text. 
The details the input of Alogorithm~\ref{alg:adaptive-annealing-slmc} are as follows:
\begin{itemize}
    \item A target distribution $p(x; \beta) \propto \tilde{p}(x; \beta)$ : A probability distribution from which a user wishes to sample, where the normalizing factor is not necessary to compute acceptance probability.
    \item An initial $\beta_{0}$ : The inverse temperature where the annealing starts, where $\beta_{0}$ should be set so that the distribution is easily explorable even with a local update MCMC method.
    \item Training data $\mathcal{D}_0$ : Training data used to train the initial model $p_{\theta_{0}}(x; \beta_{0})$. 
    \item An initial model $p_{\theta_{0}}(x; \beta_{0})$ : A model trained with the training data $\mathcal{D}_{0}$.
    \item An initial distribution $\pi_{0}$ : An initial distribution of VAE-SLMC during annealing. 
    In simulated annealing, it is necessary to simulate the $\beta_{k}$ probability distribution with the final state of the simulation of the $\beta_{k-1}$ probability distribution as the initial value $x^{(0)}_{k}$ because the transition between modes becomes more difficult as the inverse temperature is large. However, VAE-SLMC works for arbitrary initial distributions because the proposal is independent of previous states, and transitions between modes are easy. Therefore, the algorithm uses the same initial distribution for all $k$ without loss of generality.\label{line:explain-init-dist} 
\end{itemize}
In all the numerical experiments,
we set $T_{\mathrm{train}} = \infty,~~\forall k \in \{1, \ldots, K\}$ in order not to perform the retraining steps in the VAE-SLMC in from Algorithm~\ref{alg:VAE-SLMC}.

\begin{algorithm}[t]
\caption{Adaptive Annealing VAE-SLMC}\label{alg:adaptive-annealing-slmc}\begin{algorithmic}[1]
\renewcommand{\algorithmicrequire}{\textbf{Input:}}
\renewcommand{\algorithmicensure}{\textbf{Output:}}
\REQUIRE a target distribution $\tilde{p}(x;\beta)$, an initial $\beta_{0}$, training data $\mathcal{D}_{0}$, an initial model $p_{\theta_{0}}(x ; \beta_{0})$, an initial distribution $\pi_{0}$.
\STATE Initialize $x^{(0)} \sim \pi_{0}$
\FOR{$k=1$ to $K$}
    \STATE $\beta_{k} = \text{BETA-SEARCH}(\beta_{k-1}; \mathrm{AR}_{\min}, \mathrm{AR}_{\max}, \varepsilon, \{\beta_{(s)}^{\prime}\}_{s=0}^{S}, T_{\max}, \mathcal{D}_{k-1})$
    ($rhd$ See Algorithm~\ref{alg:adaptive-anneal} \label{alg:adaptive-annealing-slmc-adaptive-beta})
    \STATE Set $(x^{(t)}_{k})_{t=0}^{T} = \text{VAE-SLMC}(\tilde{p}(x ; \beta_{k}), p_{\theta_{k-1}}(x ; \beta_{k-1}), \Gamma_{\theta_{k-1}}(x; \beta_{k-1}); T, T_{\mathrm{train}}, \pi_{0}, \mathcal{D}_{k-1})$ from Algorithm 1.
    \STATE Make the training data $\mathcal{D}_{k}$ from $(x^{(t)}_{k})_{t=0}^{T}$
    \STATE Train $p_{\theta_{k}}$ with $\mathcal{D}_{k}$
\ENDFOR
\STATE Set $(x^{(t)}_{K})_{t=0}^{T} = \text{VAE-SLMC}(\tilde{p}(x ; \beta_{K}), p_{\theta_{K}}(x ; \beta_{K}), \Gamma_{\theta_{K}}(x; \beta_{K}); T, T_{\mathrm{train}}, \pi_{0}, \mathcal{D}_{K})$
\ENSURE samples $(x^{(t)}_{K})_{t=0}^{T}$
\end{algorithmic}
\end{algorithm}

In Line~\ref{alg:adaptive-annealing-slmc-adaptive-beta} of Algorithm~\ref{alg:adaptive-annealing-slmc},
$\beta$ is adaptively determined by Algorithm~\ref{alg:adaptive-anneal},
whose inputs are detailed as follows:
\begin{itemize}
    \item $\beta$ : A current value of $\beta$.
    \item $\mathrm{AR}_{\min}$, $\mathrm{AR}_{\max}$ : Upper and lower bounds for the acceptance rate of the VAE-SLMC at each annealing step. The upper and lower bounds are the values determined by a user according to the acceptance rate sought at the end of the annealing $\beta_{K}=1$. For a typical behavior of the lower and upper bounds of the adaptive annealing VAE-SLMC, please refer to Sec.~\ref{sec:dependency-lower-bound}.
    \item $\varepsilon$ : A parameter used to update the candidates $\{\beta_{(s)}^{\prime}\}_{s=0}^{S}$ in a sequential search step.
    \item $\{\beta_{(s)}^{\prime}\}_{s=0}^{S}$ : An initial sequence of $\beta$ candidates $\{\beta_{(s)}^{\prime}\}_{s=0}^{S}$.
    They consist of the input $\beta$ and smaller and larger values around $\beta$ with a user-defined granularity, e.g., an equal interval or a logarithmically equal interval.
    \item $T_{\max}$ : The maximum number of iterations for sequential $\beta$ search.
    \item $T_{\mathrm{check}}$ : The number of the Monte Carlo steps in VAE-SLMC used to estimate acceptance rates of the $\beta$ candidates $\{\beta_{(s)}^{\prime}\}_{s=0}^{S}$. 
\end{itemize}
The followings are additional notes with respect to Algorithm~\ref{alg:adaptive-anneal}:
\paragraph{Parallel calculation of acceptance rates.}
In order to calculate the acceptance rates of candidates $\{\beta_{s}^{\prime}\}_{s=0}^{S}$ in Line~\ref{alg:adaptive-anneal-para} by a local update MCMC method, it is necessary to calculate the acceptance rate by $S$ times local update MCMC simulations. Computation of the acceptance rates in Line~\ref{alg:adaptive-anneal-para} can be executed in parallel since SLMC proposes candidates independent of the previous state. In other words, the overhead of calculating the acceptance rates for $S$ candidates by using SLMC only changes the number of calculating acceptance rates from $1$ to $S$ for each step of VAE-SLMC.

\paragraph{Selection from $\beta$ candidates.}
In Line~\ref{alg:adaptive-anneal-select},
if there are multiple $\beta$s that satisfy the conditions, you can select $\beta_{(s)}^{\prime}$ according to arbitrary rules. For example, if you select the largest $\beta$ among all the $\beta$s satisfying the condition, the number of annealing steps will be small, but the acceptance rate will be about the lower bound. On the other hand, if you select the smallest $\beta$ among all the $\beta$s satisfying the conditions, the number of annealing steps will be larger, but the acceptance rate will be about the upper bound.
\begin{algorithm}[t]
\caption{Adaptive $\beta$ determination}\label{alg:adaptive-anneal}
\begin{algorithmic}[1]
\renewcommand{\algorithmicrequire}{\textbf{Input:}}
\renewcommand{\algorithmicensure}{\textbf{Output:}}
\REQUIRE a current $\beta$, lower bound $\mathrm{AR}_{\min}$, upper bound $\mathrm{AR}_{\max}$, sequential search step $\varepsilon$, candidates $\{\beta^{\prime}_{(s)}\}_{s=0}^{S}$, max iterations $T_{\max}$, Monte Carlo steps $T_{\mathrm{check}}$, training data $\mathcal{D}$ 
\FOR{$t=1$ to $T_{\max}$} 
\STATE $\rhd$ Parallel search \label{alg:adaptive-anneal-para}
\STATE Calculate $\{\mathrm{AR}(\beta_{(s)}^{\prime})\}_{s=0}^{S}$ by $\text{VAE-SLMC}(\tilde{p}(x; \beta^{\prime}_{(s)}), p_{\theta_{\beta}}(x;\beta), \Gamma_{\theta_{\beta}}(x; \beta);  T_{\mathrm{check}}, T_{\mathrm{train}}, \pi_{0}, \mathcal{D})$ 
\IF{$\exists \mathrm{AR}_{\min} \le \mathrm{AR}(\beta_{(s)}^{\prime}) \le \mathrm{AR}_{\max},~~ \forall s \in \{0, \ldots, S\}$}
\STATE Select $\beta_{(s)}^{\prime}$ from $\{\beta_{(s)}^{\prime} \mid \mathrm{AR}_{\min} \le \mathrm{AR}(\beta_{(s)}^{\prime}) \le \mathrm{AR}_{\max},~\forall s \in \{0, \ldots, S\}\}$.
 \label{alg:adaptive-anneal-select}
\STATE \textbf{break}
\STATE \textbf{Return} $\beta^{\prime}_{(s)}$
\ELSE
\STATE $\rhd$ Sequential search
\IF{$\mathrm{AR}(\beta_{s}) \le AR_{\min},~\forall s \in \{0, \ldots, S\}$}
\STATE $\beta_{(s)}^{\prime} \gets \beta^{\prime}_{(s)} - \varepsilon,~~\forall s \in \{0, \ldots, S\}$
\ELSE
\STATE $\beta_{(s)}^{\prime} \gets \beta^{\prime}_{(s)} + \varepsilon,~~\forall s \in \{0, \ldots, S\}$
\STATE
\ENDIF
\ENDIF
\ENDFOR
\ENSURE $\beta^{\prime}_{(s)}$
\end{algorithmic}
\end{algorithm}

\subsection{Parallel Annealing Exchange VAE-SLMC}\label{subsec:parallel-anneal}
Algorithm \ref{alg:parallel-anneal} illustrates the steps of parallel annealing exchange VAE-SLMC explained in  Sec.~3.4 of the main text. The details the input of Algorithm~\ref{alg:parallel-anneal} are as follows: 
\begin{itemize}
    \item A target distribution $p(x; \beta) \propto \tilde{p}(x; \beta)$ : A probability distribution from which a user wishes to sample, where the normalizing factor is not necessary to compute acceptance probability.
    \item A $\beta$-sequence $\{\beta_{k}^{j}\}_{k=0, \ldots, K}^{j=0, \ldots, J}$ : A set of $\beta$s to determine the annealing schedules for all the parallel chains.
    $j$ and $k$ indices represent the parallel chains and the annealing steps, independently.
    \item An initial model $\{p_{\theta_{0}^{j}}(x; \beta_{0}^{j})\}_{j=0}^{J}$ : A model trained with the training data $\{\mathcal{D}^{j}_{0}\}_{j=0}^{J}$ from $\{p(x;\beta_{0}^{j})\}_{j=0}^{J}$ by a local update EMC method.
    \item Training data $\{\mathcal{D}^{j}_{0}\}_{j=0}^{J}$ : Training data used to train the initial model $\{p_{\theta_{0}^{j}}(x; \beta_{0}^{j})\}_{j=0}^{J}$. 
    \item An initial distribution $\pi_{0}$ : An initial distribution of VAE-ESLMC during annealing. We use the same initial distribution in the annealing for the same reasons as described in Sec.~\ref{subsec:aa-vae-slmc}.
\end{itemize}

\begin{algorithm}[p]
\caption{Parallel Annealing VAE-ESLMC}\label{alg:parallel-anneal}
\begin{algorithmic}[1]
\renewcommand{\algorithmicrequire}{\textbf{Input:}}
\renewcommand{\algorithmicensure}{\textbf{Output:}}
\REQUIRE a target distribution $\tilde{p}(x)$, a $\beta$-sequence $\{\beta_{k}^{j}\}_{k=0, \ldots, K}^{j=0, \ldots, J}$, an initial models $\{p_{\theta_{0}^{j}}(x;\beta_{0}^{j})\}_{j=1}^{J}$, an initial distribution $\pi_{0}$, training data $\{\mathcal{D}^{j}_{0}\}_{j=0}^{J}$
\STATE Initialize $\{x^{j,(0)}\}_{j=0}^{J} \sim \pi_{0}$
\STATE Set $\{(x_{0}^{j, (t)})\}_{j=1}^{J} = \text{VAE-ESLMC}(\{\tilde{p}(x;\beta_{0}^{j})\}_{j=1}^{J}, \{p_{\theta^{j}}(x; \beta^{j}_{0})\}_{j=0}^{J}, \{\Gamma_{\theta^{j}_{0}}(x; \beta^{j}_{0})\}_{j=0}^{J}, \{\beta^{j}_{0}\}_{j=0}^{J};\newline  T, T_{\mathrm{train}}, \{\mathcal{D}^{j}_{0}\}_{j=0}^{J}, \pi_{0})$ 
($\rhd$ See Algorithm.~\ref{alg:exchange-slmc} \label{alg:parallel-annealing-slmc-vae-eslmc})
\FOR{$k=1$ to $K$}
\STATE Set $\{(x_{k}^{j, (t)})\}_{j=1}^{J} = \text{VAE-ESLMC}(\{\tilde{p}(x;\beta_{k}^{j})\}_{j=1}^{J}, \{p_{\theta^{j}}(x; \beta^{j}_{k-1})\}_{j=0}^{J}, \{\Gamma_{\theta^{j}_{k-1}}(x; \beta^{j}_{k-1})\}_{j=0}^{J}, \{\beta^{j}_{k}\}_{j=0}^{J};\newline  T, T_{\mathrm{train}}, \{\mathcal{D}^{j}_{k-1}\}_{j=0}^{J}, \pi_{0})$ 
($\rhd$ See Algorithm.~\ref{alg:exchange-slmc} \label{alg:parallel-annealing-slmc-vae-eslmc})
\STATE Make the training data $\{\mathcal{D}_{k}^{j}\}_{j=0}^{J}$ from $\{(x_{k}^{j, (t)})\}_{j=1}^{J}$.
\STATE Train $\{p_{\theta_{k}^{j}}(x;\beta_{k}^{j})\}_{j=0}^{J}$ with $\{\mathcal{D}_{k}^{j}\}_{j=0}^{J}$
\ENDFOR
\STATE Set $\{(x_{K}^{j, (t)})\}_{j=1}^{J} = \text{VAE-ESLMC}(\{\tilde{p}(x;\beta_{K}^{j})\}_{j=1}^{J}, \{p_{\theta^{j}_{K}}(x; \beta^{j}_{K})\}_{j=0}^{J}, \{\Gamma_{\theta^{j}_{K}}(x; \beta^{j}_{K})\}_{j=0}^{J}, \{\beta^{j}_{K}\}_{j=0}^{J}; \newline T, T_{\mathrm{train}}, \{\mathcal{D}_{K}^{j}\}_{j=0}^{J}, \pi_{0})$
\ENSURE samples $(x^{(t), J}_{K})_{t=0}^{T}$
\end{algorithmic}
\end{algorithm}

\begin{algorithm}[p]
\caption{VAE-ESLMC}\label{alg:exchange-slmc}
\begin{algorithmic}[1]
\renewcommand{\algorithmicrequire}{\textbf{Input:}}
\renewcommand{\algorithmicensure}{\textbf{Output:}}
\REQUIRE $\{\tilde{p}(x;\beta^{j})\}_{j=0}^{J}$, $\{p_{\theta^{j}}(x;\beta^{j})\}_{j=0}^{J}$, $\{\Gamma_{\theta^{j}}(x;\beta^{j})\}_{j=0}^{J}$, $\{\beta^{j}\}_{j=0}^{J}$, $T$, $T_{\mathrm{train}}$, $\{\mathcal{D}^{j}\}_{j=0}^{J}$, $\pi_{0}$
\STATE Initialize $x^{(0)} \sim \pi_{0}$
\WHILE{$t \le T$}
    \FOR{$j=0$ to $J$}
    \STATE Generate $x^{\prime, j} \sim p_{\theta^{j}}(x;\beta^{j})$
    \STATE Calculate $A(x^{ \prime, j}| x^{(t), j})$ by Eq.~(5) in main text
\IF{($A(x^{\prime}| x^{(t)}) > U[0,1]$)}
    \STATE $x^{(t+1), j} \gets x^{\prime, j}$ 
\ELSE
    \STATE $x^{(t+1), j} \gets x^{(t), j}$ 
\ENDIF
\IF{$t \equiv 0 ~(\mathrm{mod}~ T_{\mathrm{train}})$}
    \STATE Retrain Step
    \STATE Make training data $\mathcal{D}^{(t)}$ from $(x^{(t)})_{t=0}^{t}$ 
    \STATE Set $\mathcal{D}^{j} \gets \mathcal{D}^{j} \cup \mathcal{D}^{(t), j}$
    \STATE Train $p_{\theta^{j}}(x^{j};\beta^{j})$ with $\mathcal{D}^{j}$ and update $\Gamma_{\theta^{j}}(x^{j} ; \beta^{j})$ 
    ($\rhd$ Use the last parameters as initial parameters for training)
\ENDIF
\ENDFOR
\STATE Select $x^{(t), j} \in \{x^{(t), j}\}_{j=0}^{J}$
\STATE Calculate $A_{\mathrm{ex}}(x^{(t), j+1}, x^{(t), j})$ by Eq.~\eqref{eq:exchange-mc-prob}
\IF{$A_{\mathrm{ex}}(x^{(t), j+1}| x^{(t), j}) > U[0, 1]$}
\STATE $x^{(t), j} \gets x^{(t), j+1)}$
\STATE $x^{(t), j+1} \gets x^{(t), j}$
\ENDIF
\STATE $t \gets t+1$
\ENDWHILE
\ENSURE samples $(x^{(t), j})_{t=0, \ldots T}^{j=0, \ldots J}$
\end{algorithmic}
\end{algorithm}

In Line~\ref{alg:parallel-annealing-slmc-vae-eslmc} of Algorithm~\ref{alg:parallel-anneal}, samples are generated by VAE-ESLMC described in Algorithm~\ref{alg:exchange-slmc}, which uses the same exchange process as the replica exchange Monte Carlo method \citep{swendsen1986replica, hukushima1996exchange}. For all parallel chains, the exchange probability is defined as
\begin{equation}
\label{eq:exchange-mc-prob}
    A_{\mathrm{ex}}(x^{j+1}, x^{j}) = \min \left(1, \frac{\tilde p(x^{j+1};\beta^{j}) \tilde p(x^{j};\beta^{j+1})}{\tilde p(x^{j};\beta^{j}) \tilde p(x^{j+1};\beta^{j+1})}\right).
\end{equation}
The same ``retrain'' step as in VAE-SLMC can also be incorporated into the VAE-ESLMC\@. However, retraining is not performed in all numerical experiment (i.e., $T_{\mathrm{train}} = \infty$).

\section{ADDITIONAL RESULTS}\label{sec:additional-results}
In this section, we report additional results omitted from the main text because of limited space.

\subsection{Comparing VAEs and Flow based model}\label{subsec:vae-vs-flow}
This section compares  our proposed VAE-SLMC with other VAE-SLMC proposed in \citet{monroe2022learning} and Flow-SLMC proposed in \citet{albergo2019flow}. We briefly describe the VAE-based method proposed in \citet{monroe2022learning}. Although \citet{monroe2022learning} use the prior $p_{\theta}(z)$ and the posterior $q_{\phi}(z| x)$ constructed by Flow-based model, we consider the simple prior $p(z) = \mathcal{N}(0, I)$ and posterior $q_{\phi}(z| x) = \mathcal{N}(z ; \mu_{\phi}(x), \mathrm{diag}(\sigma^{2}_{\phi}(x)))$ case in Sec~\ref{subsec:implict-isometricity} for simplicity. The basic idea of \citet{monroe2022learning} is that a well-trained VAE may reach the case that the true posterior $p_{\theta}(z | x)$ and variational posterior $q_{\phi}(z | x)$ coincide, i.e., $p_{\theta}(z | x) \approx q_{\phi}(z | x)$, and approximates the likelihood function as
\begin{equation*}
    p_{\theta}(x) = \frac{p_{\theta}(z) p_{\theta}(x | z)}{p_{\theta}(z | x)} \approx \frac{p_{\theta}(z) p_{\theta}(x | z)}{q_{\phi}(z | x)}.
\end{equation*}

Next, we describe the Flow-based model (RealNVP) used in following comparison. We use an affine coupling layer consisting of a three-layer fully-connected neural network with leaky rectified linear activations, refered in \citet{albergo2019flow}. On the other hand, The VAE was the same VAE commonly used in the numerical experiments in this paper; see Supplementary Material \ref{subsec:VAE-condition-numerical-experiment} for the details of the VAE. 

\begin{figure}[tb]
    \centering
    \includegraphics[scale=0.5]{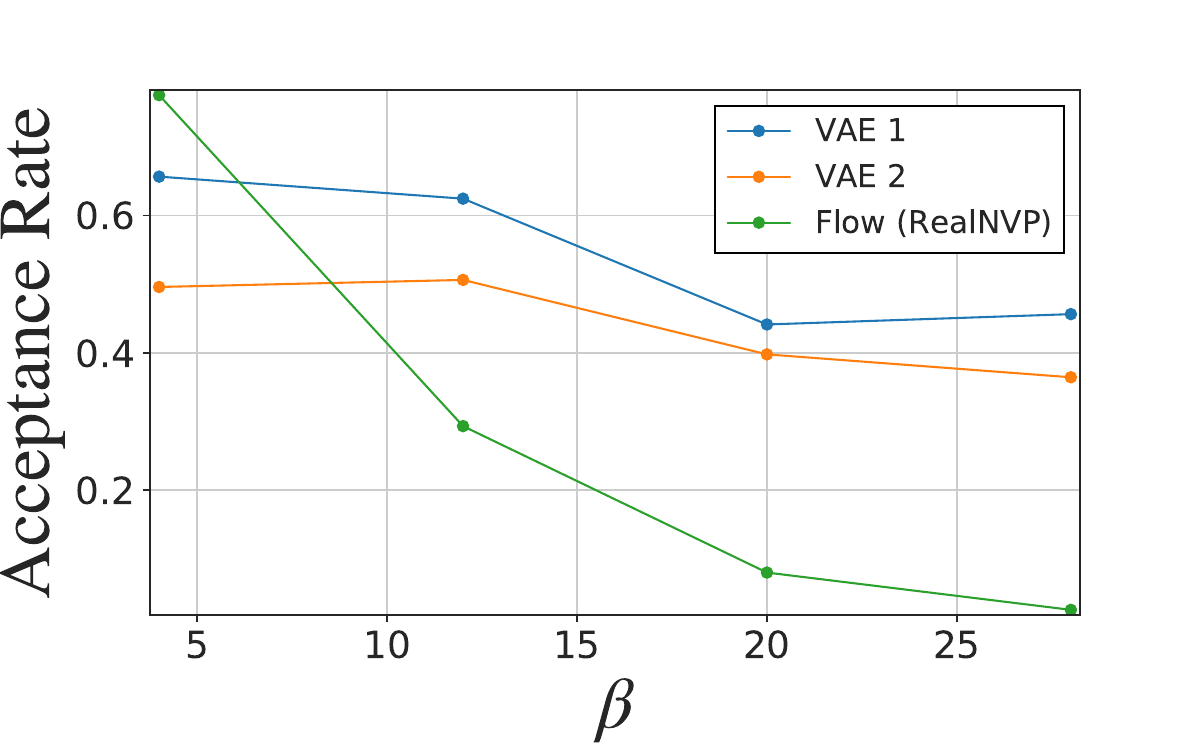}
    \caption{Comparison of our proposed VAE method, method of Flow based model in \cite{albergo2019flow} and method of VAE in \cite{monroe2022learning} by acceptance rate. Every dotted/solid line depicts the mean of acceptance rates of $10$ independent chains on annealing paths. The color represent the methods: VAE 1/VAE 2/Flow (RealNVP) are our proposed VAE method/VAE method in \cite{monroe2022learning}/flow method in \cite{albergo2019flow}. The number of parameters of the models for each dimension is summarized in the table \ref{tab:number-of-parameter}.}
    \label{fig:flow-vs-vae}
\end{figure}

The details of the number of parameters for each model are summarized in Table \ref{tab:number-of-parameter}. Each model was trained until the loss nearly converged. Figure~\ref{fig:flow-vs-vae} shows the relationship between each dimension of target distribution and the acceptance rate to compare the differences between the models and the methods (VAE 1: our proposed method, VAE 2: method in \citet{monroe2022learning}) on $5$ cluster Gaussian mixture at $\beta=1.0$. We observe that our method obtains the highest acceptance rate, even in a limited setting. An exhaustive search is in the future work.

\begin{table}[t]
\label{tab:number-of-parameter}
\caption{Number of parameters for VAE and Flow.}\label{tab:VAE-SLMC-summary}
\begin{center}
\begin{tabular}{lcccc}
\hline
Dimension  &  4  & 12 & 20 & 28 \\
Number of VAE parameters  & 263,244 & 269,924 & 276,604 & 283,284 \\
Number of Flow parameters  & 270,352 & 282,672 & 294,992 & 307,312
\end{tabular}
\end{center}
\end{table}

\subsection{Dependency of $\mathrm{AR}_{\min}$ in Algorithm~\ref{alg:adaptive-annealing-slmc}}\label{sec:dependency-lower-bound}
\begin{figure}[tb]
    \centering
    \includegraphics[width=\columnwidth]{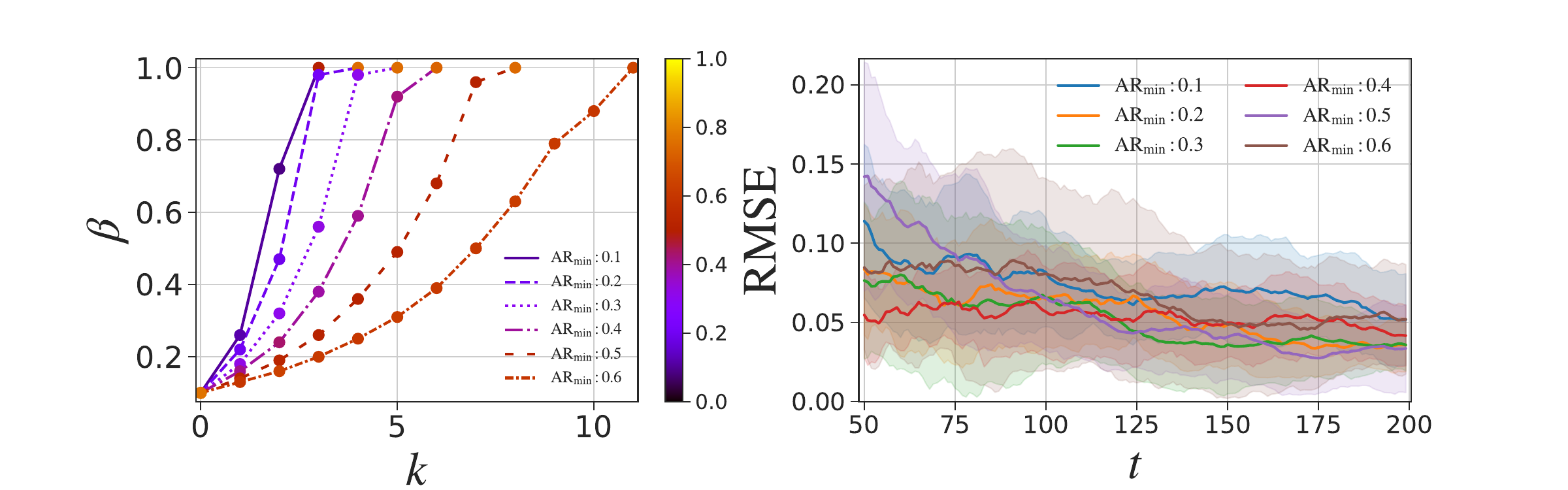}
    \caption{Dependence of the lower bound $\mathrm{AR}_{\min}$ on the $\beta$ transition, the acceptance rate, and RMSE  in a three-cluster $10$D Gaussian. (Left) The $\beta$ transition after $k$-th annealing, and acceptance rates for all $\mathrm{AR}_{\min}$. The colorbar denotes the acceptance rate. (Right) RMSE after $t$-th Monte Carlo step.}
    \label{fig:adaptive-anneal-AR-min-step}
\end{figure}

Figure~\ref{fig:adaptive-anneal-AR-min-step} shows the dependency of the lower bound $\mathrm{AR}_{\min}$ in AA-VAE-SLMC on the annealing $\beta$ and the acceptance rate denoted on the colorbar after $k$-th annealing step, and the RMSEs after $t$-th Monte Carlo step in three-cluster $10$D  Gaussian,
which are the same settings used in the main text Sec.~$4.1$ except for lower bound $\mathrm{AR}_{\min}$. 
This result confirms that the acceptance rate $\mathrm{AR}$ falls appropriately between the upper bound $\mathrm{AR}_{\max} = 1.0$ and lower bounds.

From these results, a large value of $\mathrm{AR}_{\min}$ requires a long number of annealing to reach $\beta_{K}=1$, while a small value of $\mathrm{AR}_{\min}$ results in a small number of annealing to reach $\beta_{K}=1$. Specifically, when $\mathrm{AR}_{\min} = 0.6$, 11 times of annealing ($K=11$) are required to reach $\beta_{K}=1$, but when $\mathrm{AR}_{\min}=0.1$, only 3 times of annealing ($K=3$) are required.  We also find that for any given $\mathrm{AR}_{\min}$, the RMSE decreases as same, indicating that the algorithm works for any given $\mathrm{AR}_{\min}$.
Furthermore, when $\mathrm{AR}_{\min}$ is between 0.1 and 0.5, the last beta intervals become smaller because $\beta_{K} = 1.0$, and the acceptance rates increase rapidly. These result suggests that our adaptive annealing control the acceptance rate in the middle of annealing, which is effective to set $\mathrm{AR}_{\min}$ to a low value, reduce the number of the annealing.

\subsection{Sensor Network Localization}\label{subsec:sensor-problem}
\begin{figure}[tb]
    \centering
    \includegraphics[width=\columnwidth]{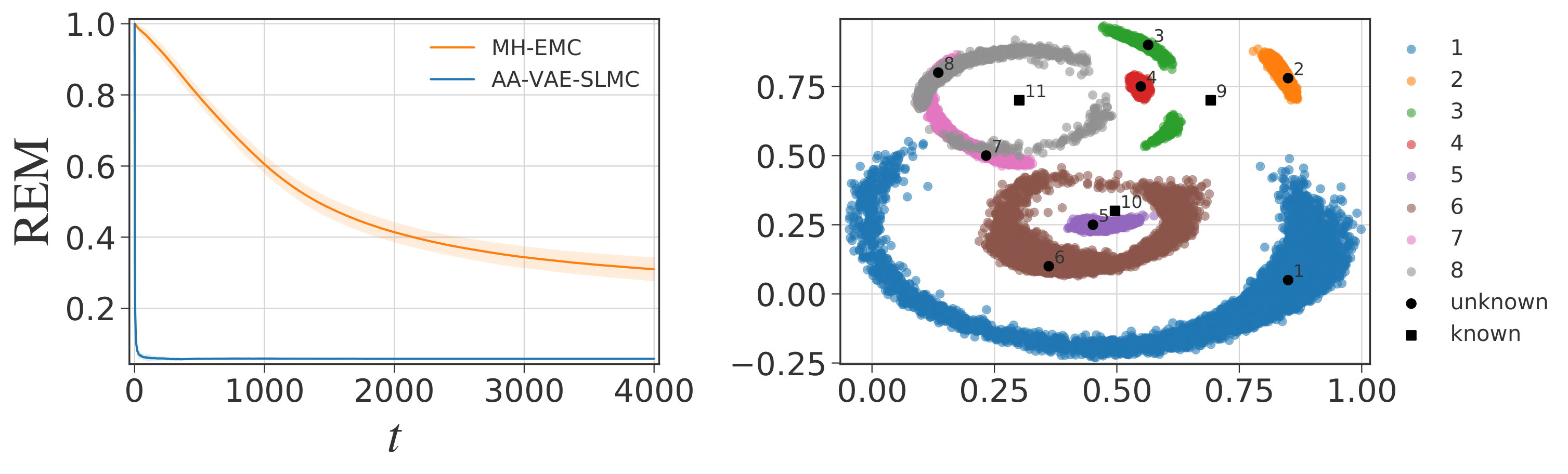}
    \caption{The relative error of the estimated mean (REM). The mean and standard deviation are computed from $10$ Markov chain at $\beta_{k} = 1.0$ with AA-VAE-SLMC. (Left) The horizontal axis indicates Monte Carlo steps. (Right) The square vertices represent the two-dimensional planar positions of known sensors, and the circle vertices represent the two-dimensional planar positions of unknown sensors. The scatter plots represent the sample sequences generated by AA-VAE-SLMC.}
    \label{fig:sensor-rem}
\end{figure}

In this section, we examine the results of AA-VAE-SLMC applied to the problem to estimate the location of a sensor network by Bayesian estimation.
This problem is typically a multimodal posterior distribution with multiple modes; see Supplementary Material Sec.~\ref{sec:experiment-detail} for details of parameter settings of the AA-VAE-SLMC\@. Specifically, we follow the experimental setup by \cite{ihler2004nonparametric} and assume that $N$ sensors have two-dimensional locations denoted by $\{x^{(s)} \in 
 \mathbb{R}^{2}\}_{s=1}^{N}$ and exist in a two-dimensional plane. The distance $(x^{(s)}, x^{(s^{\prime})})$ between a pair of sensors is observed with probability $p(x^{(s)}, x^{(s^{\prime})}) = \exp (- 0.5\|x^{(s)}-x^{(s^{\prime})}\|^{2}/R^{2})$, and the distance follows $d_{s, s^{\prime}} = \|x^{(s)} - x^{(s^{\prime})}\| + \varepsilon$, $\varepsilon \sim \mathcal{N}(0, \sigma_{\varepsilon})$ including Gaussian noise. Given a set of observations $d_{s, s^{\prime}}$ and a prior (uniform distribution) of $x^{(s)}$, a typical task is to infer all sensor positions from the posterior distribution. We set $N = 8$ ($D=16$), $R/L = 0.3$, $\sigma_{\varepsilon}/L = 0.02$, and add three sensors with known positions. The positions of eight sensors form a $16$-dimensional multimodal distribution. This distribution is difficult for a local updated MCMC method to visit all the modes, and a large bias is introduced to the obtained samples. Convergence is evaluated using the relative error of the estimated mean (REM) in all dimensions. The REM is a summary of the error in approximating the expected value of a variable across all dimensions, which is computed as
\begin{equation}
\label{eq:REM-sensor}
    \mathrm{REM}_{t} = \frac{\sum_{i=1}^{2N}|\bar{x}^{(t)}_{i} - x_{i}^{\ast} |}{\sum_{i=1}^{2N}|x_{i}^{\ast}|},
\end{equation}
where $\bar{x}^{(t)}_{i} = \sum_{\tau=1}^{t} x^{(\tau)}_{i}/\sum_{\tau=1}^{t} 1$ and denote the sampling average of the $i$-th variable at time $t$. Moreover, $x^{\ast}_{i}$ is the mean of the $i$-th variable of the target probability distribution. We first train the VAE with 40, 000 training data at $\beta_{0}=0.1$, which are generated using a local updated MCMC method with the acceptance rate tuned to almost in a range between 0.2 and 0.3. AA-VAE-SLMC start from $\beta_{0}=0.1$ and, we set $\mathrm{AR}_{\min}=0.15$, $\mathrm{AR}_{\max} =1.0$, $\varepsilon=0.01$; adaptive $\beta$ from $\beta^{\prime}=1.0$ at each $\beta^{\prime}$ proposal step. 
The left side of Fig.~\ref{fig:sensor-rem} shows the dynamics of the REM and the right side of Fig.~\ref{fig:sensor-rem} shows known and unknown sensor positions and generated samples by AA-VAE-SLMC on two-dimensional planar. 

The REM rapidly decreases for samples with a fast transition between multiple modes and small autocorrelation (see Sec.~\ref{sec:metrics}). The figure on the left shows that AA-VAE-SLMC has a much more rapidly decreasing REM than MH-EMC, which suggests that the VAE obtained by annealing is proposing between modes at high speed, producing samples with much smaller autocorrelations. The figure on the right shows that multiple modes of the posterior distribution can be sampled precisely centered on the true position of the unknown sensor.

\subsection{Optimization Problems}\label{subsec:optimization-problem}
By considering the cost function to minimize as a negative log-likelihood and annealing to sufficiently large $\beta$, the optimization problem can be solved in the same way as simulated annealing \citep{kirkpatrick1983optimization}. Here,  we demonstrate our method is extremely effective in searching for multiple globally optimal solutions. Specifically, we confirm the behavior of AA-VAE-SLMC on the following Himmelblau function \citep{himmelblau2018applied}:
\begin{equation}
\label{eq:himmelblau}
    f(x_{1}, x_{2}) = - (x_{1}^{2} + x_{2} -11)^{2} - (x_{1} + x_{2}^{2} - 7)^{2}.
\end{equation}
It is one of the famous optimization problems with the following $4$ global optima:
\begin{equation*}
(x_{1}^{\ast}, x_{2}^{\ast}) = (3.0, 2.0), (-2.805118, 3.131312),(-3.779310, -3.283186), (3.584428, -1.848126).
\end{equation*}
We use AA-VAE-SLMC method starting from $\beta_{0}=0.1$ to $\beta_{k} = 50$ and set $\mathrm{AR}_{\min} = 0.25$ and $\mathrm{AR}_{\max} = 1.0$; 

\begin{figure}[tb]
    \centering
    \includegraphics[width=\columnwidth]{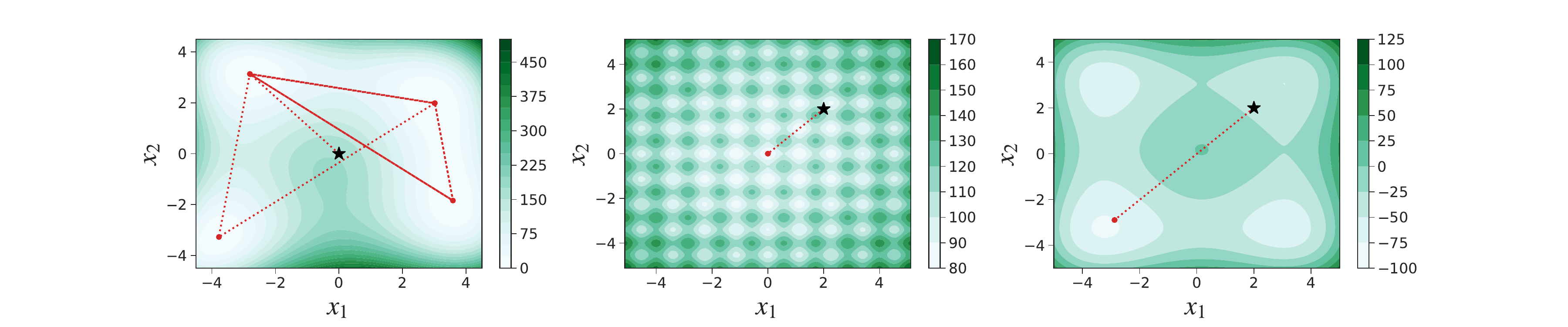}
    \caption{Snapshot of the global optimal solution search with AA-VAE-SLMC at $\beta_{K} = 50$. The red dotted line represents the first $30$ transitions. The star point is the initial value. The horizontal and vertical axis indicates the value of each dimension. The contour lines of the left, middle, and right figure represent Himmelblau (2D), Rastrigin (2D), and Styblinski-Tang (2D), respectively.}
    \label{fig:opt-transition}
\end{figure}

Figure~\ref{fig:opt-transition} (Left) shows that the generated samples consist only of states near the optimal solution and that the four optimal values are generated with almost equal probability, whose global transition is effective complex optimization problmes.
The acceptance rate is about $0.43$, and the RMSE between the estimated value of the function and the true optimal value of the function is $1.18 \times 10^{-3} \pm 4.0 \times 10^{-4}$ across 10 chain average and standard deviation up to 50,000 iterations.

We also validate the performance of AA-VAE-SLMC for other optimization problems in addition to the Himmelblau function. We use three standard benchmark test functions: Rastrigin function \citep{rastrigin1974systems}, and Styblinski--Tang function \citep{silagadze2007finding}. See Supplementary Material Sec.~\ref{sec:experiment-detail} for details of the parameter setting for AA-VAE-SLMC.

\begin{itemize}
    \item  Rastrigin function: A function that has a global optimal solution at the origin, where the function value is zero, defined by
    \begin{equation*}
        f(x) = \sum_{i=1}^{D} (x_{i}^{2} - 10 \cos (2 \pi x_{i})),
    \end{equation*}
    subject to $x_{i} \in [-5.12, 5.12]$. The global minimum is located at $x^{\ast} = (0, \ldots, 0)$, $f(x^{\ast}) = 0$.
    \item Styblinski--Tang function: A function with three local solutions and one globally optimal solution in the two-dimensional case:
    \begin{equation*}
        f(x) = \frac{1}{2} \sum_{i=1}^{D} (x_{i}^{4} - 16 x_{i}^{2} + 5 x_{i}),
    \end{equation*}
    subject to $x_{i} \in [-5, 5]$. The global minimum is located at
    \begin{equation*}
     x^{\ast} = (-2.90353\ldots, \cdots, -2.90353\ldots),~~~f(x^{\ast}) = - 39.16616 \ldots \times D.   
    \end{equation*}
\end{itemize}

Table~\ref{tab:results-optimization} shows that the RMSE between the objective function estimated by an MCMC method and the true optimal value:
\begin{equation}
\label{eq:RMSE-opt}
    \mathrm{RMSE}_{\mathrm{opt}} = \sqrt{(\bar{f}(x)-f(x^{\ast}))^{2}},
\end{equation}
where $\bar{f}(x)$ denotes the average cost function estimated by 50,000 Monte Carlo steps, under the number of annealing steps $K$, an initial beta $\beta_{0}$, and a final beta $\beta_{K}$. 
The smaller the $\mathrm{RMSE}_{\mathrm{opt}}$, the closer the generated samples are to the global optimal, meaning that the optimization problem is solved more accurately. 
As can be seen from Table~\ref{tab:results-optimization}, the RMSEa are very small in all situations, indicating that the AA-VAE-SLMC can solve the optimization problems even for problems with multiple local optimal and somewhat high dimensionality. 
Figure~\ref{fig:opt-transition} shows that snapshots of the transitions of AA-VAE-SLMC for each problem. For the problems with multiple global optimal solutions, the samples should be generated from the global optimal with almost equal probability.
In the case of a single global optimal, it is also desirable that the samples be generated in the vicinity of a single global optimal solution, skipping over the local optimal. 
As can be seen from the figure, for the Himmelblau function with four globally optimal solutions, the samples are generated on the four global optimal with almost equal probability.  
For the Rastrigin and Styblinski--Tang functions, the samples are generated immediately in the vicinity of the global optimal from the initial state. These global transitions are achieved by the information-rich latent space of well-trained VAE\@. 
Such a global transition is not feasible for simulated annealing using the local update and may be an efficient optimization method.

\begin{table}[tb]
\caption{Results for each optimization problem. RMSE of the Cost function represents the RMSE between the average cost function estimated by 50,000 MCMC for 10 chains and the true optimal solution.}\label{tab:results-optimization}
\begin{center}
\begin{tabular}{lcccc}
Target  & $\mathrm{RMSE}_{\mathrm{opt}}$ in Eq.~\eqref{eq:RMSE-opt} & Annealing steps $K$ & Initial beta $\beta_{0}$ & Final beta $\beta_{K}$ \\
\hline
Rastrigin ($2$D)        &$3.00 \times 10^{-5} \pm 1.91 \times 10^{-7}$ & $10$ & $0.1$ & $50$ \\
Rastrigin ($10$D)     &$9.74 \times 10^{-2} \pm 3.22 \times 10^{-4}$ & $16$ & $0.1$ & $50$ \\
Styblinski--Tang ($2$D)       &$1.52 \times 10^{-3} \pm 2.92 \times 10^{-6}$ & $8$ & $0.1$ & $50$ \\
Styblinski--Tang ($10$D)       &$3.58 \times 10^{-3} \pm 9.66 \times 10^{-4}$ & $15$ & $0.1$ & $50$ \\
\end{tabular}
\end{center}
\end{table}

\subsection{Experimets of Na\"ive VAE-SLMC}\label{sec:naive-vae-slmc}
In this section, we show that even the na\"ive VAE-SLMC can generate samples more efficiently than HMC, the most common method for continuous probability distributions when good training data can be obtained.
We evaluate the performance of the na\"ive VAE-SLMC  which uses only Algorithm~1, i.e., without annealing, on the following toy test distributions:
\begin{itemize}
    \item Ill-conditioned Gaussian (ICG): A 100-dimensional multivariate normal distribution with mean $0$ and variance $\Sigma$. Eigenvalues of $\Sigma$ range from $10^{-2}$ to $10^{2}$ and eigenvectors are chosen in a random orthonormal basis.
    \item Strongly correlated Gaussian (SCG): A diagonal 2-dimensional normal distribution with variance $[10^{2} , 10^{-2}]$ rotated by $\pi/4$. This is an extreme version of the example by \cite{neal2011mcmc}.
    \item Banana-shaped density (BANANA): A 2-dimensional probability distribution as
    \begin{equation*}
        x_{1} \sim \mathcal{N}(0, 10);~~~x_{2} \sim \mathcal{N}(0.03(x_{1}^{2}-100), 1).
    \end{equation*}
    This distribution has a strong ridge-like geometrical structure that is difficult for HMC~\citep{duane1987hybrid} to understand its structure.
    \item Rough well density (RW): A distribution similar to the example by \cite{sohl2014hamiltonian}, for a given $\eta > 0$,
    \begin{equation*}
        p(x) \propto \exp \left(- \frac{1}{2} x^{\top} x + \eta \sum_{i=1}^{n} \cos \left(\frac{x_{i}}{\eta} \right) \right).
    \end{equation*}
    For a small $\eta$, the distribution itself does not change much, but the surface is perturbed by high-frequency tines oscillating between $-1$ and $1$. In our numerical experiments, we set $\eta=10^{-2}$.
\end{itemize}

For each of them, we compare the na\"ive  VAE-SLMC with Hamilton Monte Carlo (HMC)~\citep{duane1987hybrid} which fix $10$ leapfrog steps and use well-tuned step size. 
We also set $n(\mathcal{D}) = 20,000$ to train a VAE\@. The samples used as training data were generated by well-tuned HMC and spaced long enough to be almost independent samples; see Supplementary Materials Sec.~\ref{sec:details-EMC-HMC-MH} for details of parameters of HMC.
In this numerical experiment, the performance of VAE-SLMC is evaluated by ESS, which is commonly used in HMC evaluations \citep{levy2017generalizing}, instead of RMSE\@. 
Specifically, we measured performance via the mean across runs of the minimum effective sample size (ESS) across dimensions and the 1st and 2nd order moments, motivated by \cite{levy2017generalizing}; see Supplementary Materials Sec.~\ref{sec:metrics} for details of definition and meaning.
Table~\ref{tab:VAE-SLMC-summary} summarizes the results for the distributions. The na\"ive  VAE-SLMC performs well in the experiments. This result indicates that MCMC can be significantly accelerated if a good sequence of samples is obtained by some method.

\begin{table}[t]
\caption{Effective sample size (ESS) for the na\"ive   VAE-SLMC and HMC. ESS depends on the statistic being estimated; we compute it for the 1st moment and 2nd moment of each dimension and report the minimum across statistics and dimensions of the mean ESS across $10$ runs for the na\"ive   VAE-SLMC and HMC, report $\pm$ standard errors; see the supplementary material for the behaviors of the autocorrelation function.}\label{tab:VAE-SLMC-summary}
\begin{center}
\begin{tabular}{lcc}
Target  & HMC  & Na\"ive  VAE-SLMC\\
\hline
Ill-conditioned Gaussian ($100$D)     &$2.58 \times 10^{-2} \pm 8.02 \times 10^{-4}$ & $5.20 \times 10^{-1} \pm 9.46 \times 10^{-3}$ \\
Strongly correlated Gaussian ($2$D)   &$1.50 \times 10^{-2} \pm 7.11 \times 10^{-4}$ & $9.65 \times 10^{-1} \pm 7.62 \times 10^{-3}$\\
Banana-shaped density ($2$D)    &$9.32 \times 10^{-2} \pm 1.66 \times 10^{-3}$ &$2.30 \times 10^{-1} \pm 3.69 \times 10^{-3}$\\
Rough well density ($2$D)  &$5.22 \times 10^{-4} \pm 1.96 \times 10^{-4}$ &$3.89 \times 10^{-3} \pm 9.24 \times 10^{-4}$\\
\end{tabular}
\end{center}
\end{table}

\section{EXPERIMENT DETAILS}\label{sec:experiment-detail}
\subsection{Model Details}
The followings are additional details of the models used in the experiments of the main paper.

\paragraph{Gaussian Mixture.}
In Sec.~4.1 of the main text, we first define $i_{D} = (i, \ldots, i) \in \mathbb{R}^{D}$, $\sigma^{2} = 0.5 \sqrt{D/100}$ and consider the following Gaussian mixture models:
\begin{itemize}
    \item $2$-cluster gaussian mixture:
    \begin{equation*}
    p(x) = \frac{1}{2} \mathcal{N}(- 1_{D}, \sigma^{2} I_{D}) + \frac{1}{2} \mathcal{N}(1_{D}, \sigma^{2} I_{D})
    \end{equation*}
    \item $3$-cluster Gaussian mixture:
    \begin{equation*}
    p(x) = \frac{1}{3} \mathcal{N}(- 2_{D}, \sigma^{2} I_{D}) + \frac{1}{3} \mathcal{N}(0_{D}, \sigma^{2} I_{D}) 
    + \frac{1}{3} \mathcal{N}(2_{D}, \sigma^{2} I_{D})
    \end{equation*}
    \item $4$-cluster Gaussian mixture:
    \begin{equation*}
    p(x) = \frac{1}{4} \mathcal{N}(- 3_{D}, \sigma^{2} I_{D}) + \frac{1}{4} \mathcal{N}(- 1_{D}, \sigma^{2} I_{D}) + \frac{1}{4} \mathcal{N}(1_{D}, \sigma^{2} I_{D}) \\
    + \frac{1}{4} \mathcal{N}(3_{D}, \sigma^{2} I_{D})
    \end{equation*}
    \item $5$-cluster Gaussian mixture:
    \begin{equation*}
    p(x) = \frac{1}{5} \mathcal{N}(- 4_{D}, \sigma^{2} I_{D}) + \frac{1}{5} \mathcal{N}(- 2_{D}, \sigma^{2} I_{D}) + \frac{1}{5} \mathcal{N}(0_{D}, \sigma^{2} I_{D}) \\
    + \frac{1}{5} \mathcal{N}(2_{D}, \sigma^{2} I_{D}) + \frac{1}{5} \mathcal{N}(4_{D}, \sigma^{2} I_{D})
    \end{equation*}
\end{itemize}

\paragraph{Spectral Analysis.}
In Sec.~4.2 of the main text, we obtain the data $\{y(kT)\}_{k=0}^{K-1}$ by discretizing $y(\tau)$ with a period $T$:
\begin{equation*}
    y(kT) = A_{0} + \sum_{i=1}^{N} A_{i} \cos(2 \pi f_{i} k T + \phi_{i}) + r(kT),
\end{equation*}
where $r(kT) \sim \mathcal{N}(0, \sigma^{2})$, and  we define the following posterior to estimate frequencies $f$:
\begin{equation*}
    p(f) \propto \exp \left(\frac{1}{2\sigma^{2}} \left( y(kT) - A_{0} - \sum_{i=1}^{N} A_{i} \cos (2 \pi f_{i} T k + \phi_{i})  \right)^{2} \right) \mathbf{1}[f],
\end{equation*} 
where $\mathbf{1}[f]$ denotes a indicator function such that $\mathbf{1}[f] = 1$ if $f \in [0, 0.5]^{N}$ and $\mathbf{1}[f] = 0$ if $f \notin [0, 0.5]^{N}$, which is based on the periodicity outside of $[0, 0.5]^{N}$. 
Furthermore, for the sake of simplicity, let assume that $T$ and $\sigma$ are known, we set $A_{0}=0$, $A_{i} = 1, \forall i =1, \ldots, N$, $f = (0.1, 0.4, \ldots, 0.1, 0.4) \in \mathbb{R}^{8}$.

\subsection{Experimental Conditions for the Annealing VAE-SLMC Methods}
We describe the specific hyperparameters of the annealing of CA-VAE-SLMC, AA-VAE-SLMC, CA-VAE-ESLMC, and AA-VAE-ESLMC for each experiment in Table~\ref{tab:annealing-setting}. 
The algorithms for each method explained in Table~\ref{tab:ablation-alg} are invoked with these hyperparameters.
In all the numerical experiments, retrain step in VAE-SLMC (Algorithm~1) and VAE-ESLMC (Algorithm~\ref{alg:exchange-slmc}) was not performed for each annealing (i.e., $T_{\mathrm{train}}=\infty, ~~\forall k = 1, \ldots, K$). 
We did not perform parallel search in Algorithm \ref{alg:adaptive-anneal} for simplicity (i.e., $S=1$), and sequential search was performed from $\beta_{s}^{\prime}=1$ to input $\beta$. $T_{\max}$ was set to the value obtained by dividing the interval between $\beta_{s}^{\prime}=1$ and input $\beta$ by $\varepsilon$.
\begin{table}[tb]
\caption{The annealing conditions for each experiment. The abbreviations and algorithms for each ``Method'' are summarized in Table.\ref{tab:ablation-alg}.}\label{tab:annealing-setting}
\begin{center}
\begin{tabular}{lcccccccc}
Experiment & Method & $\beta_{0}$ & $\mathrm{AR}_{\min}$ & $\mathrm{AR}_{\max}$ & $\varepsilon$ & $T_{\mathrm{check}}$ & Epoch & $n(\mathcal{D}_{k})$ \\
\hline
Gaussian Mixture & CA-VAE-SLMC & $0.1$ &  &  &  & $2, 000$ & $150$ & $15,000$ \\
& AA-VAE-SLMC & $0.1$ & $0.1$ & $1.0$ & $0.01$  & $2, 000$ & $150$ & $15,000$ \\
& AA-VAE-ESLMC & $0.1$ & $0.2$ & $1.0$ & $0.01$ & $2, 000$ & $150$ & $15,000$ \\
Spectral Analysis & AA-VAE-SLMC & $0.1$ & $0.2$ & $1.0$ & $0.01$ & $2, 000$ & $150$ & $15,000$ \\
Sensor Problems & AA-VAE-SLMC & $0.05$ & $0.2$ & $1.0$ & $0.01$  & $2, 000$ & $150$ & $15,000$ \\
Optimization Problems & AA-VAE-SLMC & $0.1$ & $0.2$ & $1.0$ & $0.01$ & $2, 000$ & $150$ & $15,000$ \\
\end{tabular}
\end{center}
\end{table}

\subsection{Conditions for Training the VAE in VAE-SLMC}\label{subsec:VAE-condition-numerical-experiment}
In this section, we summarize the parameter settings for model assignment and learning for VAE.
\paragraph{VAE configulations.}
We define $\mathrm{FC}(i, o, f)$ as a fully-connected layer with input dimension $i$, output dimension $o$, and activate function $f$, and $D$ and $M$ as a input dimension and a latent dimension, respectively. We used the same network and set $M=D$ for all numerical experiments in Sec.~4 and Sec.~\ref{sec:additional-results}. The encoder is composed of $\mathrm{FC}(\mathrm{D}, 256, \mathrm{gelu})$-$\mathrm{FC}(256, 256, \mathrm{gelu})$-$\mathrm{FC}(256, 128, \mathrm{gelu})$-$\mathrm{FC}(128, 128, \mathrm{gelu})$-$\mathrm{FC}(128, 64, \mathrm{gelu})$-$\mathrm{FC}(64, 32, \mathrm{gelu})$-$\mathrm{FC}(32, \mathrm{M}, \mathrm{gelu})$ and the decoder is $\mathrm{FC}(\mathrm{M}, 32, \mathrm{gelu})$-$\mathrm{FC}(32, 64, \mathrm{gelu})$-$\mathrm{FC}(64, 128, \mathrm{gelu})$-$\mathrm{FC}(128, 128, \mathrm{gelu})$-$\mathrm{FC}(128, 256, \mathrm{gelu})$-$\mathrm{FC}(256, 256, \mathrm{gelu})$-$\mathrm{FC}(256, \mathrm{D}, \mathrm{gelu})$.

\paragraph{Training hyperparameters.}
Table~\ref{tab:learning-setting} summarizes the training parameters other than the network. In all numerical experiments, we use an Adam optimizer and the sum of squared errors (SSE) for the objective function.

\begin{table}[tb]
\caption{The training hyper parameters on each experiment for all $k=1, \ldots, K$ in Algorithms~2, \ref{alg:adaptive-annealing-slmc}, and \ref{alg:parallel-anneal}.}\label{tab:learning-setting}
\begin{center}
\begin{tabular}{lcccccccc}
Experiment &  Learning rate & Batch size & $\beta_{\mathrm{VAE}}$  & Input dimension D & Latent dimension M \\
\hline
Gaussian Mixture  & $1.0 \times 10^{-3}$ & $516$ & $1/120$  & $[2, 10, 20, 25, 50, 100]$ & $[2, 10, 20, 25, 50, 100]$\\
ILG   & $1.0 \times 10^{-3}$ & $516$ & $6$  & $100$ & $100$\\
SG   & $1.0 \times 10^{-3}$ & $516$ & $6$  & $2$ & $2$\\
BANANA   & $1.0 \times 10^{-3}$ & $516$ & $1/20$  & $2$ & $2$\\
RW   & $1.0 \times 10^{-3}$ & $516$ & $1/20$  & $2$ & $2$\\
Himmelblau   & $1.0 \times 10^{-3}$ & $516$ & $1/200$  & $2$ & $2$\\
rastrigin    & $1.0 \times 10^{-3}$ & $1024$ & $1/300$  & $10$ & $10$\\
rastrigin    & $1.0 \times 10^{-3}$ & $1024$ & $1/300$  & $2$ & $2$\\
Styblinskitank    & $1.0 \times 10^{-3}$ & $516$ & $1/30$  & $10$ & $10$\\
Styblinskitank   & $1.0 \times 10^{-3}$ & $516$ & $1/200$  & $2$ & $2$\\
Sensor problem   & $1.0 \times 10^{-3}$ & $516$ & $1/600$  & $16$ & $16$\\
Spectral Analysis   & $1.0 \times 10^{-3}$ & $516$ & $1/1000$  & $8$ & $8$\\
\end{tabular}
\end{center}
\end{table}

\subsection{Details of the Replica Exchange Monte Carlo (EMC) and Metropolis--Hastings (MH) Methods in Table~1.}\label{sec:details-EMC-HMC-MH}
In this section, we show each method and its parameters used in the experiments. For the Metropolis--Hastings (MH) method, we used the proposal with Gaussian distribution $q(x^{\prime} | x^{(t)}) = \mathcal{N}(x^{\prime} | x^{(t)}, \sigma^{2} I)$ and adjusted the scale parameter $\sigma^{2}$ so that the acceptance rate is between 0.15 and 0.25. In the replica exchange Monte Carlo (EMC) method, the interval of the inverse temperatures is set so that the exchange acceptance rate between replicas is about 0.3 with equal probability by following \cite{nagata2008asymptotic}. When the kernel of each replica is $q(x^{\prime} | x^{(t)}) = \mathcal{N}(x^{\prime} | x^{(t)}, \sigma^{2} I)$, we adjusted scale parameter $\sigma^{2}$ so that the acceptance rate is between 0.15 and 0.25.
When the kernel of each replica is Hamilton Monte Carlo (HMC) kernel, the number of steps is fixed to 5, and the step size is set so that the acceptance rate is about between 0.4 and  0.6. The EMC method with HMC was implemented using the TensorFlow Probability's \texttt{tfp.mcmc.ReplicaExchangeMC}\footnote{\url{https://www.tensorflow.org/probability/api_docs/python/tfp/mcmc/ReplicaExchangeMC}} implementation.

\subsection{Detail of Metrics}
\label{sec:metrics}
This section provides the details and purposes of the three metrics used in our experiments.

\paragraph{Root-mean-square error (RMSE).}
RMSE is defined as the Euclidean distance between the true expected value and its estimate:
\begin{equation}
    \mathrm{RMSE}_{t} = \sqrt{\frac{1}{D} \sum_{d=1}^{D} (\bar{x}_{d}^{(t)} - x_{d}^{\ast})^{2}},
\end{equation}
where $\bar{x}^{(t)}_{d} = \sum_{t=1}^{t} x^{(t)}_{d}/\sum_{t=1}^{t} 1$ denotes the sampling average of the $d$-th dimension of the variable at Monte Carlo step $t$, and $x^{\ast}_{d}$ is the mean of the $d$-th dimension of the variable mathematically derived from the target distribution. A rapid decrease of $\mathrm{RMSE}_{t}$ means a faster convergence to a mean of the target probability distribution. When all modes of the target distribution cannot be accurately explored, RMSE does not converge to zero.

RMSE is helpful to evaluate whether the samples are biased or not. 
Furthermore, the convergence rate takes longer if the samples cannot efficiently move between modes. Therefore, we used RMSE in Gaussian mixture experiments in main text Sec.~4.1 to ensure that annealing VAE-SLMC efficiently moves between modes and generates unbiased samples.

\paragraph{Relative error (REM).}
A summary of the error in approximating the expected value of a variable across all dimensions computed by
\begin{equation}
\label{eq:REM}
    \mathrm{REM}_{t} = \frac{\sum_{d=1}^{D}|\bar{x}^{(t)}_{d} - x_{d}^{\ast} |}{\sum_{d=1}^{D}|x_{d}^{\ast}|},
\end{equation}
where $\bar{x}^{(t)}_{d}$ and $x^{\ast}_{d}$ are the same as in RMSE above.
REM is divided by the true mean $x^{\ast}_{d}$ to eliminate dependence on the scale for each dimension $d$. A rapid decrease of $\mathrm{REM}_{t}$ means a faster convergence to a mean of the target probability distribution.

REM is also helpful to evaluate the bias of samples like RMSE.
Since REM is scale-invariant by considering the normalization for each dimensional component of sample space, we used REM for the spectral problem in main text Sec.~4.2 and sensor problems in Sec.~\ref{subsec:sensor-problem} where  the scales of each dimension are different.

\paragraph{Effective sample size (ESS).}
ESS is a common measure to evaluate efficiency of sampling algorithms. The ESS represents the estimated number of iid samples generated. When measuring ESS in higher dimensions, it is common to consider ESS for each dimension. For a sequence of samples $(x^{(t)})_{t=0}^{T}$, the effective sample size (ESS) was computed as 
\begin{equation}
    \mathrm{ESS} = \frac{1}{1 + 2 \sum_{t=1}^{\hat{T}} \hat{\rho}_{t}},
\end{equation}
where $\hat{T}$ is the stopping time. $\hat{\rho}_{t}$ is an estimate of the lag-$t$ autocorrelations of the Markov chain:
\begin{equation}
    \hat{\rho}_{t} = \frac{1}{N-t} \sum_{r=1}^{N-t} (x^{(\tau)}-\bar{x})(x^{(\tau + t)} - \bar{x}).
\end{equation}
The ESS is calculated by using TensorFlow Probability's \texttt{tfp.mcmc.effective\_sample\_size}\footnote{\url{https://www.tensorflow.org/probability/api_docs/python/tfp/mcmc/effective_sample_size}} implementation. If an MCMC simulation generates completely independent samples, then $\mathrm{ESS}=1$. We report the mean across 10 runs of minimum ESS across dimensions and the 1st and the 2nd moments.

ESS is a helpful measure to quantify the correlation between samples and to see the number of samples that are effectively independent and is the most commonly used evaluation measure in Bayesian estimation for gradient-based MCMCs such as HMC\@. On the other hand, it is difficult for ESS to capture whether the transition between modes is efficient visually. Therefore, in the performance evaluation of Na\"ive VAE-SLMC in Sec.~\ref{sec:naive-vae-slmc}, we use ESS as the metric because it is not a particularly multimodal distribution, and we wanted to confirm its superiority over HMC for distribution families that HMC is not good at.


\end{document}